\documentclass[acmsmall]{acmart}
\usepackage{array}
\usepackage{longtable}
\usepackage{graphicx}
\usepackage{makecell}
\usepackage{booktabs}
\usepackage{float}
\AtBeginDocument{%
  }

\setcopyright{acmlicensed}
\copyrightyear{2024}
\acmYear{2024}
\acmDOI{XXXXXXX.XXXXXXX}

\begin{document}

\title{A Survey on Kolmogorov-Arnold Network}

%% "authornote" and "authornotemark" commands.

\author{Shriyank Somvanshi}
\affiliation{%
  \institution{Texas State University}
  \city{San Marcos}
  \country{TX}}
\email{jum6@txstate.edu}

\author{Syed Aaqib Javed}
\affiliation{%
  \institution{Texas State University}
  \city{San Marcos}
  \country{TX}}
\email{aaqib.ce@txstate.edu}

\author{Md Monzurul Islam}
\affiliation{%
  \institution{Texas State University}
  \city{San Marcos}
  \country{TX}}
\email{monzurul@txstate.edu}

\author{Diwas Pandit}
\affiliation{%
  \institution{Texas State University}
  \city{San Marcos}
  \country{TX}}
\email{zxh15@txstate.edu}

\author{Subasish Das, Ph.D.}
\affiliation{%
  \institution{Texas State University}
  \city{San Marcos}
  \country{TX}}
\email{subasish@txstate.edu}

\renewcommand{\shortauthors}{Somvanshi et al.}

\begin{abstract}
  This systematic review explores the theoretical foundations, evolution, applications, and future potential of Kolmogorov-Arnold Networks (KAN), a neural network model inspired by the Kolmogorov-Arnold representation theorem. KANs set themselves apart from traditional neural networks by employing learnable, spline-parameterized functions rather than fixed activation functions, allowing for flexible and interpretable representations of high-dimensional functions. The review delves into KAN’s architectural strengths, including adaptive edge-based activation functions that enhance parameter efficiency and scalability across varied applications such as time series forecasting, computational biomedicine, and graph learning. Key advancements—including Temporal-KAN (T-KAN), FastKAN, and Partial Differential Equation (PDE) KAN illustrate KAN’s growing applicability in dynamic environments, significantly improving interpretability, computational efficiency, and adaptability for complex function approximation tasks. Moreover, the paper discusses KAN’s integration with other architectures, such as convolutional, recurrent, and transformer-based models, showcasing its versatility in complementing established neural networks for tasks that require hybrid approaches. Despite its strengths, KAN faces computational challenges in high-dimensional and noisy data settings, sparking continued research into optimization strategies, regularization techniques, and hybrid models. This paper highlights KAN’s expanding role in modern neural architectures and outlines future directions to enhance its computational efficiency, interpretability, and scalability in data-intensive applications.
\end{abstract}



%\ccsdesc[300]{Do Not Use This Code~Generate the Correct Terms for Your Paper}
%\ccsdesc{Do Not Use This Code~Generate the Correct Terms for Your Paper}
%\ccsdesc[100]{Do Not Use This Code~Generate the Correct Terms for Your Paper}


\received{08 November 2024}

\maketitle

\section{Introduction}
Kolmogorov-Arnold Networks (KAN) are a class of neural networks inspired by the Kolmogorov-Arnold representation theorem, which posits that any multivariate continuous function can be expressed as a sum of continuous functions of one variable. Developed by Andrey Kolmogorov and Vladimir Arnold \cite{kolmogorov1956representation}, this theorem provides a foundational understanding of how high-dimensional functions can be decomposed into simpler, univariate components, which has inspired the creation of KANs as a novel neural architecture. Rather than employing traditional fixed activation functions, KANs utilize learnable, spline-parametrized univariate functions on edges, allowing for a more adaptive function representation \cite{liu2024kan}. While KANs are less widely adopted than more conventional models such as Convolutional Neural Networks (CNN) or Recurrent Neural Networks (RNNs), their mathematical underpinnings offer a strong theoretical framework for tasks involving high-dimensional function approximation.

KANs leverage this insight by replacing traditional neural network weights with learnable univariate functions, enabling a more flexible and interpretable framework for function approximation. This structural shift differentiates KANs from Multi-Layer Perceptrons (MLPs), which use fixed activation functions at the nodes, offering KANs the advantage of adaptability and greater alignment with the decomposition of multivariate functions \cite{schmidt2021kolmogorov}. This architecture has gained attention in machine learning as a potentially more parameter-efficient and theoretically grounded alternative to traditional deep learning models. KANs have demonstrated superior performance in applications such as predicting flexible electrohydrodynamic pump parameters, where they provide both accuracy and interpretability through symbolic formula extraction \cite{peng2024predictive}. As KANs continue to evolve, variations such as the Chebyshev KAN, which enhances nonlinear function approximation through Chebyshev polynomials, are emerging as promising developments in the field \cite{ss2024chebyshev}.

Recent research on Kolmogorov-Arnold Networks (KANs) has demonstrated their potential as efficient and interpretable alternatives to traditional MLPs \cite{samadi2024smooth}. KANs differ from MLPs by replacing linear weights with learnable activation functions, enabling dynamic pattern learning and improved performance with fewer parameters \cite{vaca2024kolmogorov}. The growing interest in KANs stems from their ability to achieve comparable or even superior accuracy to larger MLPs, faster neural scaling laws, and enhanced interpretability.

However, notable gaps in the current literature persist. One major gap involves KANs' limitations in efficiently representing smooth, high-dimensional functions. Finite KAN structures often struggle with exact function approximation, resulting in challenges with training convergence and their applicability to complex, real-world data \cite{samadi2024smooth}. Further, questions arise about the robustness of KANs when applied to diverse datasets, especially in comparison to more established deep learning architectures such as Long Short-Term Memory Networks (LSTMs) and CNNs \cite{vaca2024kolmogorov}. While recent advancements like Wavelet KAN (Wav-KAN) have sought to improve both interpretability and computational efficiency, additional research is necessary to optimize the interaction between wavelets and KANs for large-scale data applications \cite{bozorgasl2024wav}. The continued interest in KANs is largely driven by their potential to reduce the number of learnable parameters while enhancing both accuracy and interpretability, particularly in fields that require data-efficient models and high-level explanations.

Since their inception, KANs have undergone significant evolution, advancing both in theoretical foundations and practical applications. Initially proposed as an alternative to MLPs, KANs leverage the Kolmogorov-Arnold representation theorem to approximate multivariate functions using univariate functions, providing enhanced interpretability and parameter efficiency \cite{samadi2024smooth}. Over time, several modifications, such as the introduction of smooth KANs \cite{vaca2024kolmogorov} and spline-based activation functions \cite{peng2024predictive}, have improved KANs’ ability to capture complex nonlinearities. KANs have further evolved with architectures like Temporal Kolmogorov-Arnold Networks (TKAN), which incorporate memory management for sequential data, demonstrating superior performance over RNNs in time series forecasting \cite{genet2024tkan}. Additionally, KANs have demonstrated notable computational efficiency compared to traditional architectures like MLPs, often requiring fewer parameters while maintaining accuracy \cite{peng2024predictive, ss2024chebyshev}. These advancements underscore KANs' strengths in interpretability and efficiency, although their scalability and performance relative to more advanced architectures like CNNs and Transformers require further exploration.

In more recent developments, KANs have continued to expand across both theoretical and practical domains. Building upon the Kolmogorov-Arnold theorem, KANs have been integrated into neural cognitive diagnosis models to enhance interpretability without sacrificing performance \cite{yang2024endowing}. The introduction of FastKAN by Li \cite{li2024kolmogorov}, which approximates B-splines using Gaussian radial basis functions (RBF), significantly improved KANs' computational efficiency, making them more viable for real-world applications. Comparative studies highlight KANs' ability to reduce the number of parameters while achieving performance on par with CNNs and RNNs, as evidenced in tasks like image classification \cite{bodner2024convolutional}. Moreover, KANs have been shown to outperform transformers in tasks involving smaller datasets, delivering competitive accuracy with lower computational costs \cite{jamali2024learn}. These advancements position KANs as a scalable and efficient alternative to more complex architectures, particularly in environments constrained by data or computational resources.

The practical application of KANs has also advanced significantly, particularly with the adoption of edge-based activations, which differ from traditional networks that position activation functions at the nodes. This edge-based structure enhances KANs’ modularity and interpretability \cite{moradi2024kolmogorov}. KANs have been successfully integrated into a range of neural architectures, such as autoencoders and time series models, and have demonstrated competitive performance against CNNs, RNNs, and transformers in tasks like image reconstruction and multivariate time series forecasting, effectively capturing complex dependencies \cite{inzirillo2024sigkan}. Although KANs generally require fewer parameters than MLPs and CNNs and offer superior computational efficiency \cite{wang2024kolmogorov}, they may still fall short in tasks involving highly complex geometries, where traditional architectures retain certain advantages.

\subsection{Research Questions}

This review seeks to address several key research questions regarding KANs:

\begin{enumerate}
    \item \textbf{What are the primary theoretical developments in KANs, and how do they contribute to the broader landscape of neural network architectures?}
    
    This question aims to explore how the Kolmogorov-Arnold representation theorem has influenced the design of KANs and what theoretical innovations have emerged over time.
    
    \item \textbf{How have KANs been applied across various fields?}
    
    By addressing this, the review examines the practical applications of KANs and compares their performance with traditional architectures such as CNNs, RNNs, and transformers.
    
    \item \textbf{What are the key challenges and opportunities for KAN research, particularly in terms of scalability, computational efficiency, and robustness?}
    
    This question focuses on the limitations KANs face in large-scale applications and complex datasets, identifying potential areas for future research and optimization.
\end{enumerate}

\begin{figure}[htp]
    \centering
    \includegraphics[width=0.98\linewidth]{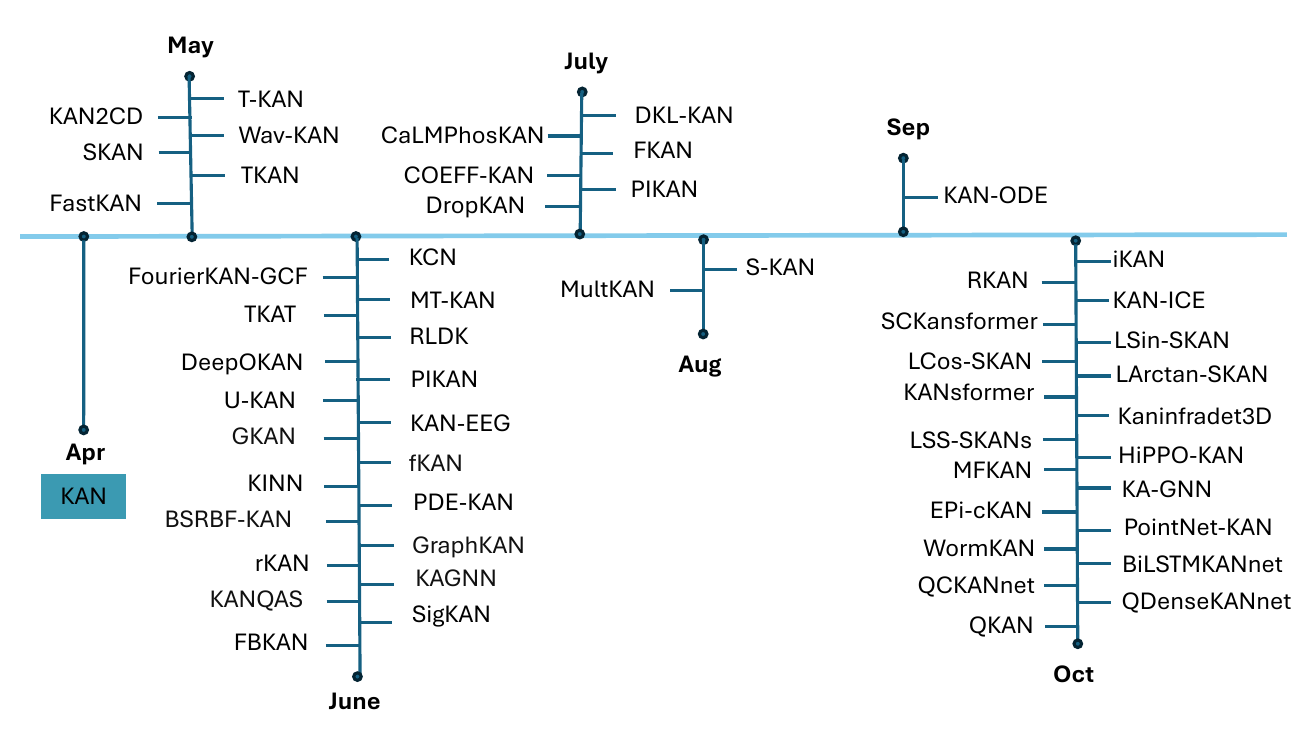}
    \caption{Progression of KAN's (2024)}
    \label{fig:progression}
\end{figure}

\section{Historical Evolution of KAN}
\subsection{Early Research}
The Kolmogorov-Arnold theorem has significantly influenced the development of KANs by providing the theoretical foundation for representing continuous multivariate functions as compositions of simpler univariate functions. Kolmogorov’s foundational work demonstrated that any continuous multi-variable function can be broken down into a sum of single-variable functions, making it possible to simplify the representation of complex, high-dimensional functions \cite{kolmogorov1956representation}. This principle directly supports the structure of KAN, which uses this approach to handle complex data more efficiently. Arnold then contributed a practical refinement, showing that functions with three variables could be represented with even fewer components, enhancing the efficiency and applicability of the theorem \cite{kolmogorov1956representation}. Together, their contributions laid the groundwork for KAN’s ability to approximate complex functions with greater interpretability and accuracy. 
This principle is reflected in KANs, which use finite network topologies to approximate complex functions. One notable advantage of KANs over traditional MLPs is their ability to place learnable activation functions on the edges of the network rather than at the nodes, which enhances flexibility and parameter efficiency. Some Studies emphasize the advantages of KANs in tasks requiring smooth function approximations, such as computational biomedicine \cite{samadi2024smooth, liu2024kan}. These networks, which utilize splines as univariate functions, demonstrate higher accuracy and interpretability compared to traditional networks, making them valuable for scientific applications. Additionally, the integration of Chebyshev polynomials into KANs improves approximation accuracy and convergence, especially for nonlinear functions, further enhancing their relevance in modern tasks that require precise nonlinear approximations \cite{ss2024chebyshev}. Figure \ref{fig:progression} provides a timeline of these advancements in KAN architectures, showcasing key developments like T-KAN, Wav-KAN, and FastKAN. These milestones illustrate the steady evolution of KAN models, each contributing to greater computational efficiency, scalability, and applicability across diverse fields.

Despite challenges such as maintaining smoothness and ensuring efficient convergence, KANs remain highly relevant due to their ability to leverage structural system knowledge, improving both data efficiency and model interpretability. The continuing importance of the Kolmogorov-Arnold theorem in neural network architectures is underscored by its influence on the design of modern deep learning models, such as ReLU networks. Modifications to the original theorem have allowed KANs to align more closely with contemporary deep learning practices, making them more effective for complex function approximation \cite{schmidt2021kolmogorov}. The adaptability of KANs is further highlighted by their performance in time series forecasting tasks, where KANs, with their dynamic spline-based univariate functions, require fewer parameters than MLPs to model complex patterns, demonstrating their continued efficiency and interpretability \cite{vaca2024kolmogorov}.

Further developments have expanded the foundational capabilities of KANs, with hybrid models like TKANs combining the strengths of KANs with memory mechanisms from RNNs and LSTM networks. TKANs' effectiveness in handling long-term dependencies in sequential data, outperforming traditional models in multi-step forecasting tasks, demonstrates the continued relevance of the Kolmogorov-Arnold theorem in cutting-edge neural network design \citeN{genet2024tkan}. KANs' flexibility is further demonstrated by the integration of gated mechanisms similar to LSTM and GRU cells, which enable efficient learning without extensive regularization \cite{selitskiy2022kolmogorov}. This dynamic architecture allows KANs to adjust activation functions according to task complexity, emphasizing the theorem's role in optimizing neural network efficiency. In practical applications, KANs continue to excel, as demonstrated by their ability to outperform Random Forest and MLP models in predicting pressure and flow rates in electrohydrodynamic pumps, achieving lower mean squared error (MSE) while providing interpretable symbolic formulas \cite{peng2024predictive}. However, the practical application of the Kolmogorov-Arnold theorem can be limited by the non-smoothness of inner functions, complicating network construction \cite{girosi1989representation}. Despite this critique, the theorem remains essential for understanding the properties of modern neural networks.

Advancements in KANs, such as the Wav-KAN described by Bozorgasl and Chen \cite{bozorgasl2024wav}, leverage wavelet transforms to improve both performance and interpretability, particularly for tasks requiring multi-resolution analysis, showcasing KANs' broad applicability in modern fields like signal processing. Similarly, the influence of the Kolmogorov-Arnold theorem on the design of deep ReLU networks is highlighted, as these networks approximate continuous functions through superposition, mitigating the curse of dimensionality while maintaining computational efficiency \cite{montanelli2020error}. Efforts to enhance the speed and efficiency of KAN implementations, such as FastKAN, introduce Gaussian RBFs to approximate B-splines, boosting computational speed without sacrificing accuracy \cite{li2024kolmogorov}. These innovations underscore the ongoing practical significance of the Kolmogorov-Arnold theorem in neural networks, particularly for tasks demanding efficient and accurate modeling. Funahashi's work \cite{funahashi1989approximate} further reinforces the theorem's impact on understanding multilayer networks' ability to approximate continuous functions, influencing the development of neural network architectures and solidifying the theorem's foundational role in modern neural network design and applications.

\subsection{Kolmogorov-Arnold Theorem}
The Kolmogorov–Arnold Network (KAN) is a foundational structure in neural network theory, demonstrating that any continuous multivariate function can be represented as a sum of univariate functions \cite{liu2024kan, sun2024evaluating}. Since the 1980s and 1990s, KAN has been recognized for its potential to simplify high-dimensional mappings through interpretable, layered architectures \cite{le2024exploring,liu2024kan}. Early studies focused on translating the univariate decomposition theorem into practical neural architectures, emphasizing mathematical rigor and functional versatility for machine learning applications.

Initial implementations of KAN-based networks were limited to elementary function approximation tasks within low-dimensional spaces due to computational constraints of the time \cite{drokin2024kolmogorov}. Researchers explored neural architectures that adopted KAN's decomposition concept to validate its feasibility in neural networks, providing a foundation for later, more advanced designs \cite{toscano2024inferring,toscano2024inferring}.

A significant challenge in adapting this network model for broader applications has been balancing computational efficiency with the univariate decomposition structure \cite{ta2024fc}. While KAN offered a systematic approach to function representation, its architectural demands on high-dimensional data introduced substantial computational overhead \cite{alter2024robustness}. Issues related to memory use and processing speed became increasingly apparent, prompting innovative solutions to enhance scalability without compromising interpretability \cite{nagai2024kolmogorov,wang2024cest}.

Recent advancements have extended KAN from its theoretical foundation, introducing models like Function Combinations in KAN, which incorporate splines and radial basis functions to achieve greater representational flexibility \cite{vaca2024kolmogorov,jamali2024learn}. These extensions broaden KAN's applicability across areas such as molecular dynamics, graph neural networks (GNNs), and physics-informed simulations, though the trade-off between interpretability and computational demand persists \cite{shen2024reduced,bresson2024kagnns}. Techniques such as kernel filtering and oversampling have helped reduce some computational constraints, though they underscore the resource intensity involved in complex tasks \cite{de2024kolmogorov}.

The influence of KAN is now evident in a range of applications, including image classification with Kolmogorov-Arnold Convolutions and PDE solving in Kolmogorov-Arnold-Informed Neural Networks (KINNs). Such applications expand the network's utility to multi-scale scientific and engineering phenomena \cite{toscano2024inferring,wang2024kolmogorov}, illustrating its foundational role in neural network design. However, they also highlight the computational challenges of maintaining interpretability and performance in high-dimensional spaces \cite{kilani2024kolmogorov,liu2024kan,wang2024cest}.

KAN-based models have demonstrated potential in specialized domains such as molecular and fluid dynamics. For example, KAN-inspired interatomic potential models have improved simulation accuracy in molecular applications; however, effectively managing spline activations across large-scale systems presents a significant challenge \cite{nagai2024kolmogorov}. Similarly, in fluid dynamics, a physics-informed KAN variant, the Chebyshev Physics-Informed Kolmogorov-Arnold Network (cPIKAN or cKAN) has been applied to infer temperature fields from sparse velocity data, exhibiting effective function decomposition while facing challenges in balancing governing equations with empirical data boundaries \cite{toscano2024inferring}. The performance of KAN models declines in noisy environments as noise disrupts function approximation, requiring additional methods like kernel filtering and oversampling—though these come at the cost of increased computational demands \cite{shen2024reduced}.

For high-dimensional data processing, adaptations such as TKAN and DEEPOKAN are promising by replacing traditional neural descriptors with functions derived from KAN principles. This approach enhances interpretability, but introduces challenges with memory efficiency and computational requirements, especially with complex data types like satellite imagery and hyperspectral data \cite{vaca2024kolmogorov,jamali2024learn}. Despite its theoretical robustness, KAN-based models continue to encounter scaling challenges, particularly in noisy environments. Additional computational techniques, though effective, often complicate resource management, emphasizing the difficulties inherent in applying KAN to high-dimensional tasks \cite{shen2024reduced}. Collectively, these studies suggest that while KAN provides a reliable mathematical foundation for function approximation, achieving scalability and efficiency across diverse applications remains an ongoing challenge \cite{kilani2024kolmogorov}. Consequently, research continues to focus on reconciling interpretability with computational efficiency to enable practical neural network applications based on this influential framework.

Rapid advancements and a steady expansion across various fields have marked the progression of KAN models. Table \ref{tab:timeline_Apr_2024} and Table \ref{tab:timeline_July_2024} offer a timeline of these developments, detailing each model’s unique architecture, training methods, and defining features from April to October 2024. This chronological overview highlights the key milestones achieved in KAN research, such as improved spline-based activations, enhanced memory management for sequential data, and increased adaptability to graph structures.

%%%Table 1
\thispagestyle{empty}
\begingroup
\fontsize{8pt}{8pt}\selectfont
\begin{longtable}{p{0.15\linewidth} p{0.2\linewidth} p{0.18\linewidth} p{0.3\linewidth}}
\caption{Timeline of KAN-based Models (April- June 2024)}
\label{tab:timeline_Apr_2024} \\
\toprule
\textbf{\makecell{Model \\ (Year) \\ Source}} & 
\textbf{Architecture} & 
\textbf{\makecell{Training \\ Process}} & 
\textbf{Main Features} \\
\midrule
\endfirsthead
\toprule
\textbf{\makecell{Model \\ (Year) \\ Source}} & 
\textbf{Architecture} & 
\textbf{\makecell{Training \\ Process}} & 
\textbf{Main Features} \\
\midrule
\endhead
\midrule
\endfoot
\bottomrule
\endlastfoot

\makecell[l]{KAN \\ Apr. 2024 \\ \cite{liu2024kan}} & Learnable spline-based edge activations & Gradient descent with LBFGS, adaptive grid & High interpretability, efficient scaling laws \\
\midrule
\makecell[l]{T-KAN \\ May 2024 \\ \cite{xu2024kolmogorov}} & Spline-based, learnable edge activations & Sliding window, gradient descent, pruning & High interpretability, concept drift detection \\
\midrule
\makecell[l]{SKAN \\ May 2024 \\ \cite{samadi2024smooth}} & Structured, smooth nested functions & RMSE minimization, data-efficient, extrapolation in sparse data regions & High interpretability, scalable, effective in sparse data, smooth function representation \\
\midrule
\makecell[l]{Wav-KAN \\ May 2024 \\ \cite{bozorgasl2024wav}} & Wavelet-based, learnable edge functions & Batch norm, AdamW optimizer, grid search for wavelets & High interpretability, multi-resolution analysis, noise robustness, efficient for high-dimensional data \\
\midrule
\makecell[l]{TKANs \\ May 2024 \\ \cite{genet2024tkan}} & RKAN layers with LSTM gating, B-spline activations & Adam optimizer, RMSE loss, early stopping & High accuracy in long-term forecasting, stable training, effective memory management for sequential data \\
\midrule
\makecell[l]{KAN2CD \\ May 2024 \\ \cite{yang2024endowing}} & Two-level KANs with learnable embeddings & Adam optimizer, B-spline KANs for efficiency & High interpretability, efficient training, competitive accuracy, suitable for cognitive diagnosis \\
\midrule
\makecell[l]{FastKAN \\ May 2024 \\ \cite{li2024kolmogorov}} & Gaussian RBFs with layer normalization & Benchmarked on MNIST, 20 epochs & 3.3x faster than KAN, simplified implementation, retains accuracy for high-dimensional functions \\
\midrule
\makecell[l]{C-KAN \\ June 2024 \\ \cite{bodner2024convolutional}} & Spline-based, learnable convolutions & Gradient descent with regularization, grid updates & High efficiency, adaptable activations, competitive accuracy \\
\midrule
\makecell[l]{MT-KAN \\ June 2024 \\ \cite{xu2024kolmogorov}} & Spline-based, learnable edge activations with cross-variable interactions & Gradient descent, pruning for efficiency & High interpretability, improved multivariate forecasting \\
\midrule
\makecell[l]{SigKAN \\ June 2024 \\ \cite{inzirillo2024sigkan}} & Gated Residual KAN, path signature layer & Adam optimizer, early stopping & Accurate short-term forecasting, stable, captures temporal dependencies in complex time series \\
\midrule
\makecell[l]{GraphKAN \\ June 2024 \\ \cite{zhang2024graphkan}} & Spline-based, learnable edge activations & Cosine Annealing, LayerNorm, 200 epochs & High accuracy, effective for few-shot classification \\
\midrule
\makecell[l]{GKAN \\ June 2024 \\ \cite{kiamari2024gkan}} & Learnable spline functions on edges & Semi-supervised with backpropagation & Efficient, accurate, adjustable parameters for large-scale graph data \\
\midrule
\makecell[l]{KINN \\ June 2024 \\ \cite{wang2024kolmogorov}} & Spline-based, B-spline activations & Gradient descent, meshless sampling, triangular integration & High interpretability, efficient for PDEs, handles multi-frequency components, low spectral bias \\
\midrule
\makecell[l]{Rational KAN \\ June 2024 \\ \cite{aghaei2024rkan}} & Rational basis functions (Padé, Jacobi) & Gradient descent (L-BFGS, Adam) & High accuracy, effective for physics-informed tasks and complex approximations \\
\midrule
\makecell[l]{PDE-KAN \\ June 2024 \\ \cite{wang2024kolmogorov}} & Spline-based, B-spline activations with tanh normalization & Meshless sampling, triangular integration & Low spectral bias, efficient for multi-frequency, adaptable to complex PDEs \\
\midrule
\makecell[l]{KAGNNs \\ June 2024 \\ \cite{bresson2024kagnns}} & KAN-based, spline-driven updates & Gradient descent with early stopping & High interpretability, strong expressiveness, suited for graph regression and classification \\
\midrule
\makecell[l]{PIKANs \\ June 2024 \\ \cite{shukla2024comprehensive}} & Single-layer KAN, polynomial activations (e.g., Chebyshev) & Adam optimizer, RBA for loss adjustment & High accuracy with fewer parameters, adjustable polynomial order, enhanced stability with double precision \\
\midrule
\makecell[l]{DeepOKANs \\ June 2024 \\ \cite{shukla2024comprehensive}} & Branch [128, 100, 100, 100], Trunk [4, 100, 100, 100], Chebyshev polynomials & Adam optimizer, 200k iterations, L2 regularization & Robust to noise, strong in complex tasks, higher computational cost \\
\midrule
\makecell[l]{RLDK \\ June 2024 \\ \cite{nehma2024leveraging}} & Spline-based, compact architecture with learnable edge activations & LBFGS optimizer with custom loss (reconstruction and prediction) & High parameter efficiency, fast training, suitable for real-time control, data-efficient \\
\midrule
\makecell[l]{TKAT \\ June 2024 \\ \cite{genet2024temporal}} & Encoder-decoder with TKAN layers, self-attention & Adam optimizer, MSE loss, early stopping & High interpretability, captures temporal dependencies, suited for multivariate time series \\
\midrule
\makecell[l]{KAN-EEG \\ June 2024 \\ \cite{herbozo2024kan}} & Spline-based, learnable edge activations & 100 epochs on EEG data, with gradient descent and epoch-based convergence & High interpretability, efficient, adaptable across datasets, suitable for on-device deployment \\
\midrule
\makecell[l]{KANQAS \\ June 2024 \\ \cite{kundu2024kanqas}} & Spline-based, learnable activations in Double Deep Q-Network (DDQN) & Gradient descent with RL in DDQN & High interpretability, parameter-efficient, effective for quantum state prep and quantum chemistry \\
\midrule
\makecell[l]{KCN \\ June 2024 \\ \cite{cheon2024kolmogorov}} & Spline-based edge activations & Gradient descent with backpropagation; layer freezing & High accuracy, parameter-efficient, adaptable to complex data \\
\midrule
\makecell[l]{U-KAN \\ June 2024 \\ \cite{li2024u}} & Encoder-decoder with tokenized KAN layers & Cross-entropy \& Dice loss for segmentation; MSE for generation & High interpretability, efficient, adaptable for segmentation and generative tasks, improved accuracy \\
\midrule
\makecell[l]{FourierKAN-GCF \\ June 2024 \\ \cite{xu2024fourierkan}} & Fourier-based GCN with Fourier KAN replacing MLP & BPR loss, grid search, message \& node dropout & Efficient, strong interaction representation, robust, adaptable, easier training than spline-KAN \\
\midrule
\makecell[l]{FBKANs \\ June 2024 \\ \cite{howard2024finite}} & Domain decomposition, spline-based local KANs per subdomain & Combined data-driven and physics-informed loss, adaptive grids, parallel training & Scalable, noise-robust, compatible with enhancements \\
\midrule
\makecell[l]{BSRBF-KAN \\ June 2024 \\ \cite{ta2024bsrbf}} & Combines B-splines and Gaussian RBFs & 15 epochs (MNIST), 25 (Fashion-MNIST), AdamW optimizer & High accuracy, fast convergence, adaptable activations \\
\midrule
\makecell[l]{fKAN \\ June 2024 \\ \cite{aghaei2024fkan}} & Fractional Jacobi Neural Block (fJNB) with trainable $\alpha, \beta, \gamma$ & Adam and L-BFGS optimizers & High adaptability, improved accuracy, suitable for deep learning and physics-informed tasks \\
\midrule

\end{longtable}
\endgroup

%%%%
\subsection{Key Milestones}
The KAN framework has seen substantial theoretical and applied advancements, cementing its versatility in various machine learning fields due to its unique spline-based activation functions. Early innovations, such as Selectable KANs (S-KAN), introduced a dynamic choice of activation functions that optimized adaptability for complex tasks, including image classification and function fitting \cite{yang2024activation}. Expanding this adaptability, Graph-based Kolmogorov-Arnold Networks (GKANs) were developed to improve semi-supervised node classification through flexible information flow across nodes, outperforming traditional Graph Convolutional Networks (GCNs) in tests on the Cora dataset \cite{kiamari2024gkan}. The accuracy and interpretability of GKANs in biomedical and financial graph analysis, where model transparency is crucial, were further demonstrated in 2024 by Carlo et al \cite{de2024kolmogorov}. Graph Isomorphism Network (KAGIN) and Kolmogorov-Arnold Graph Convolution Network (KAGCN), two KAN-based architectures that replaced MLPs in graph learning tasks, were introduced to enhance model interpretability and performance in regression \cite{bresson2024kagnns}.
KANs have also excelled in real-world tabular data, where they demonstrated competitive accuracy on complex datasets, making them viable alternatives to conventional neural networks like MLPs, despite their higher computational cost \cite{poeta2024benchmarking}. The practical limitations of KANs, noting their increased resource demands and latency issues, which affect efficiency in hardware-based implementations \cite{le2024exploring}. Yet, KANs continue to be valuable in scenarios requiring high adaptability and interpretability. In computer vision, KAN-Mixer was presented as utilizing adaptive spline-based transformations to achieve competitive accuracy on datasets like MNIST and CIFAR-10, often matching the performance of more complex models like ResNet-18 \cite{cheon2024demonstrating}. Similarly, the KANICE model combined KANs with CNNs, demonstrating enhanced robustness against adversarial attacks and improved spatial pattern recognition \cite{ferdaus2024kanice}.

In digital forensics, KANs were combined with MLPs to accurately distinguish AI-generated images from real ones, making them valuable tools for digital verification \cite{anon2024detecting}. The framework’s time-series forecasting capabilities were also expanded by C-KAN, which utilized convolutional layers to capture temporal patterns in volatile datasets, such as cryptocurrency, achieving notable success in finance \cite{livieris2024c}. KANs were applied to meteorology with the Global Forecast System to improve real-time, localized wind predictions, proving particularly beneficial in complex terrains like airports \cite{alves2024use}. In healthcare, KANs have shown resilience in time-series classification, with models demonstrating robustness against adversarial attacks \cite{dong2024kolmogorov}. In transportation safety, KANs proved useful in driver monitoring systems, achieving high accuracy in identifying drivers' mobile phone usage \cite{hollosi2024detection}. Furthermore, KAN 2.0 incorporated symbolic tools like kanpiler and MultKAN, bridging AI with traditional physics by modeling physical laws, thus enhancing KANs' interpretability in scientific applications \cite{liu2024kan}.

Studies continued to refine KAN-based models for graph and networked data. S-ConvKAN improved KANs' performance by enabling efficient operation within convolutional layers, thereby expanding their applicability in image processing and function fitting \cite{yang2024activation}. Additionally, KAGIN and KAGCN were observed to enhance model transparency and expressiveness in challenging graph regression tasks \cite{bresson2024kagnns}. In driver safety monitoring, custom KAN networks were able to identify specific driver actions, such as mobile phone use, proving KANs’ adaptability to AI-based safety systems \cite{hollosi2024detection}.

%%%TABLE2
\thispagestyle{empty}
\begingroup
\fontsize{8pt}{8pt}\selectfont
\begin{longtable}{p{0.15\linewidth} p{0.2\linewidth} p{0.18\linewidth} p{0.3\linewidth}}
\caption{Timeline of KAN-based Models (July-Oct. 2024)}
\label{tab:timeline_July_2024} \\
\toprule
\textbf{\makecell{Model \\ (Year) \\ Source}} & 
\textbf{Architecture} & 
\textbf{\makecell{Training \\ Process}} & 
\textbf{Main Features} \\
\midrule
\endfirsthead
\toprule
\textbf{\makecell{Model \\ (Year) \\ Source}} & 
\textbf{Architecture} & 
\textbf{\makecell{Training \\ Process}} & 
\textbf{Main Features} \\
\midrule
\endhead
\midrule
\endfoot
\bottomrule
\endlastfoot

\makecell[l]{COEFF-KANs \\ July 2024 \\ \cite{li2024coeff}} & MoLFormer for chemical embeddings, followed by KAN layers & Fine-tuning with AdamW optimizer & High interpretability, data-efficient learning, complex relationship modeling, enhanced by CIDO method \\
\midrule
\makecell[l]{DKL-KAN \\ July 2024 \\ \cite{zinage2024dkl}} & Three-layer KAN with spline activation; KISS-GP and SKIP for scalability & Normalization, Adam optimizer, 2500 epochs on GPU & Captures discontinuities, excels on small datasets, reliable uncertainty estimates in sparse data \\
\midrule
\makecell[l]{CaLMPhosKAN \\ July 2024 \\ \cite{pratyush2024calmphoskan}} & Fused codon \& amino acid embeddings; ConvBiGRU; Wavelet-based KAN (DoG wavelet) & Binary cross-entropy loss, Adam optimizer, 10-fold cross-validation & High accuracy for phosphorylation site prediction, effective for disordered regions, rich feature representation \\
\midrule
\makecell[l]{DropKAN \\ July 2024 \\ \cite{altarabichi2024dropkan}} & Spline-based, post-activation masking & Adam optimizer, 2000 steps & Improved generalization, prevents co-adaptation, flexible scaling \\
\midrule
\makecell[l]{PIKANs \\ July 2024 \\ \cite{rigas2024adaptive}} & Adaptive KAN with spline-based activations & Adaptive gradient descent with grid updates & High accuracy, efficient PDE solver, customizable basis functions \\
\midrule
\makecell[l]{FKANs \\ July 2024 \\ \cite{zeydan2024f}} & Spline-based, learnable activations & Federated averaging, local training & High interpretability, privacy-preserving, fast convergence, stable performance \\
\midrule
\makecell[l]{S-KAN \\ Aug. 2024 \\ \cite{yang2024activation}} & Adaptive multi-activation nodes & Full training, selective, pruning & Flexible activation, robust fitting, improved generalization \\
\midrule
\makecell[l]{MultKAN \\ Aug. 2024 \\ \cite{liu2024kan20kolmogorovarnoldnetworks}} & KAN layers with addition \& multiplication nodes & Gradient descent, sparse regularization & High interpretability, modularity, handles multiplicative structures \\
\midrule
\makecell[l]{KAN-ODE \\ Sep 2024 \\ \cite{koenig2024kan}} & RBF-based, learnable Swish activations & Gradient descent, adjoint method, pruning & High interpretability, efficient on sparse data, symbolic learning \\
\midrule
\makecell[l]{KANICE \\ Oct 2024 \\ \cite{ferdaus2024kanice}} & ICBs with 3x3 \& 5x5 convolutions, KANLinear spline layers & 25 epochs on image datasets, batch normalization & Adaptive feature extraction, universal approximation, efficient in KANICE-mini \\
\midrule
\makecell[l]{RKAN \\ Oct 2024 \\ \cite{yu2024residual}} & Chebyshev polynomial-based KAN convolutions & SGD for small datasets, AdamW for large datasets & Improved feature representation, robust gradient flow, adaptable to CNNs, computationally efficient \\
\midrule
\makecell[l]{iKAN \\ Oct 2024 \\ \cite{liu2024ikan}} & Multi-encoder, KAN-based classifier, feature redistribution layer & Two-step: encoder and frozen KAN-based classifier training & High incremental learning performance, reduces catastrophic forgetting, supports heterogeneous data, uses local plasticity \\
\midrule
\makecell[l]{SCKansformer \\ Oct 2024 \\ \cite{chen2024sckansformer}} & KAN-based Kansformer, SCConv Encoder, GLAE & Adam optimizer, data augmentation, 100 epochs & High interpretability, reduced redundancy, effective global-local feature capture, fine-grained classification \\
\midrule
\makecell[l]{LSin-SKAN \\ Oct 2024 \\ \cite{chen2024larctan}} & Sine-based, single parameter & Gradient descent, stable, fast convergence & Efficient, high performance among SKAN variants \\
\midrule
\makecell[l]{LCos-SKAN \\ Oct 2024 \\ \cite{chen2024larctan}} & Cosine-based, single parameter & Gradient descent, moderate convergence, oscillates & Efficient, slightly lower accuracy than LArctan-SKAN \\
\midrule
\makecell[l]{LArctan-SKAN \\ Oct 2024 \\ \cite{chen2024larctan}} & Arctangent-based, internal scaling & Gradient descent, stable, fast convergence, high accuracy & Best accuracy, highly efficient, stable training \\
\midrule
\makecell[l]{KANsformer \\ Oct 2024 \\ \cite{xie2024kansformer}} & Transformer encoder, KAN decoder with splines & Unsupervised, Adam & Scalable, real-time inference, interpretable, supports transfer learning \\
\midrule
\makecell[l]{Kaninfradet3D \\ Oct 2024 \\ \cite{liu2024kaninfradet3d}} & KAN layers, cross-attention, KANvtransform, ConKANfuser & AdamW optimizer, staged training, Cosine Annealing & Enhanced feature fusion, high-dimensional data handling, improved 3D detection accuracy \\
\midrule
\makecell[l]{LSS-SKANs \\ Oct 2024 \\ \cite{chen2024lss}} & Single-parameterized shifted Softplus & Adam optimizer, fine-tuned learning rate & High efficiency, superior accuracy, strong interpretability, suitable for MNIST and similar tasks \\
\midrule
\makecell[l]{HiPPO-KAN \\ Oct 2024 \\ \cite{lee2024hippo}} & HiPPO-encoded, spline-based KAN & Gradient descent, MSE loss & Parameter-efficient, captures long dependencies, reduced lagging \\
\midrule
\makecell[l]{MFKAN \\ Oct 2024 \\ \cite{howard2024multifidelity}} & Low-fidelity, linear, and nonlinear KAN blocks & Pretrained low-fidelity; multifidelity training & Efficient with sparse high-fidelity data, adaptive linear and nonlinear modeling, data efficiency \\
\midrule
\makecell[l]{KA-GNN \\ Oct 2024 \\ \cite{li2024ka}} & Fourier-based, learnable edge activations, 5Å cut-off for bonds & Cross-entropy loss, Fourier message passing & High interpretability, parameter-efficient, robust molecular modeling \\
\midrule
\makecell[l]{EPi-cKAN \\ Oct 2024 \\ \cite{mostajeran2024epi}} & Chebyshev polynomial-based, interconnected sub-networks & Physics-informed MSE loss; step-decay learning rate & Accurate stress-strain predictions; combines physics and data-driven insights; efficient parameter use \\
\midrule
\makecell[l]{PointNet-KAN \\ Oct 2024 \\ \cite{kashefi2024pointnet}} & Shared KAN layers, Jacobi polynomials, permutation invariance & Adam optimizer, batch norm, max pooling & Competitive, efficient for 3D point cloud classification and segmentation \\
\midrule
\makecell[l]{WormKAN \\ Oct 2024 \\ \cite{xu2024kan}} & KAN-based encoder-decoder & Reconstruction loss, regularization, smoothness & High interpretability, concept drift detection \\
\midrule
\makecell[l]{BiLSTMKANnet \\ Oct 2024 \\ \cite{jahin2024kacq}} & BiLSTM + DenseKAN layers & 10-fold CV, GridSearchCV & Temporal dependency capture, interpretability, adaptable to sequence data \\
\midrule
\makecell[l]{QCKANnet \\ Oct 2024 \\ \cite{jahin2024kacq}} & Conv1DKAN + Quantum layers & Keras tuner, cross-validation & Efficient pattern recognition, quantum-classical hybrid, scalable \\
\midrule
\makecell[l]{QDenseKANnet \\ Oct 2024 \\ \cite{jahin2024kacq}} & DenseKAN layers + Quantum circuits & Hyperband tuning, 10-fold CV & Nonlinear modeling, high accuracy, robust for complex data \\
\midrule
\makecell[l]{QKAN \\ Oct 2024 \\ \cite{ivashkov2024qkan}} & Block-encoded layers with Chebyshev activations & Quantum circuits, parameterized learning & Handles high-dimensional data, efficient for complex approximations \\
\end{longtable}
\endgroup

%%%%

\section{Core Theoretical Concepts}
\subsection{KAN Architecture}
KANs represent a novel approach in neural network design, inspired by the Kolmogorov-Arnold Representation Theorem \cite{kolmogorov1961representation}. The key feature of KAN is its ability to replace traditional fixed linear weights with learnable univariate functions. This innovation allows KANs to efficiently model complex nonlinear functions, leading to improvements in both accuracy and interpretability.
\begin{figure}[h]
  \centering
  \includegraphics[width=\linewidth]{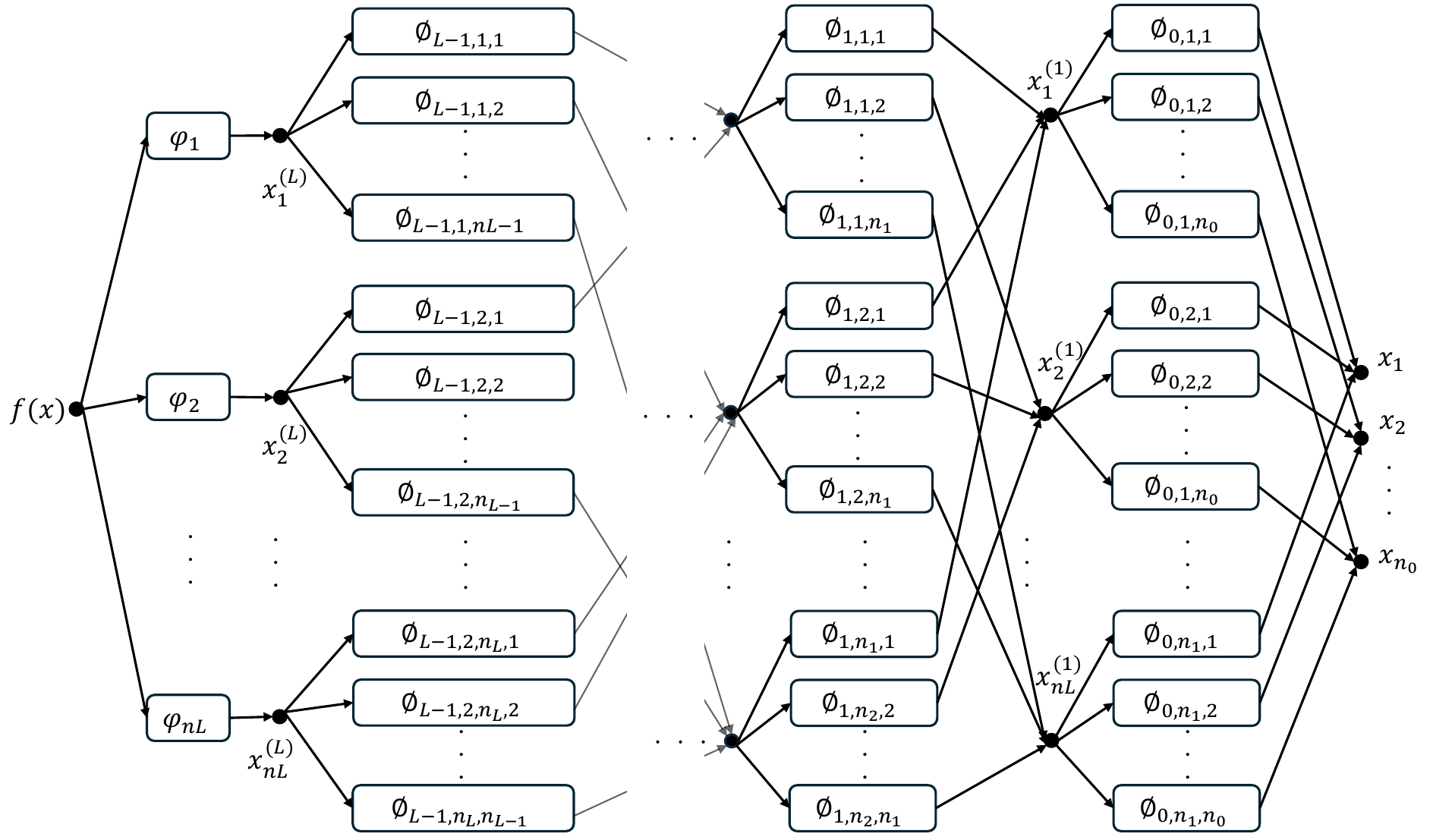}
  \caption{The Schematic of Kolmogorov-Arnold Network \cite{nagai2024kolmogorov}}
  \label{fig:schematic_of_kan}
\end{figure}
KANs are grounded in the Kolmogorov-Arnold Representation Theorem proposed by the Russian mathematician Andrey Kolmogorov in 1957, which asserts that any continuous multivariable function can be expressed as a finite superposition of continuous univariate functions \( f(x_1, x_2, x_3, \dots, x_n) \). The mathematical formulation for KANs is as follows:

\begin{equation}
f(x_1, x_2, x_3, \dots, x_n) = \sum_{q=1}^{2n+1} \phi_q \left( \sum_{p=1}^{n} \varphi_{q,p}(x_p) \right)
\label{eq:kan_function}
\end{equation}

Here, each \( \phi_q: \mathbb{R} \to \mathbb{R} \) and \( \varphi_{q,p}: [0,1] \to \mathbb{R} \) are continuous univariate functions that map input variables into a result through a sum of simpler functions. The term \( \sum_{p=1}^{n} \varphi_{q,p}(x_p) \) represents how the univariate functions are combined \cite{genet2024tkan,liu2024kan}.

All multivariate dependencies in the original function can be captured through additive combinations of univariate functions. However, to extend this to complex systems in practice, KANs use B-splines to parameterize these univariate functions, making them learnable. B-splines, which are piecewise polynomial functions, offer smooth transitions between intervals and provide local adaptability, allowing the model to fine-tune different regions of the input space independently. Unlike traditional fixed activation functions like ReLU, which apply uniform non-linearity across all nodes, KAN leverages learnable splines on edges (weights), enabling dynamic adjustments during training. This allows the model to capture intricate, non-linear relationships while maintaining overall smoothness. The learnable nature of these splines gives KAN the ability to approximate even complex, non-smooth functions that are difficult to capture with fixed activation networks. By optimizing the shape and control points of the splines during training, KAN effectively balances the representation of both global and local patterns in the data, making it more flexible, interpretable, and efficient for high-dimensional function approximation tasks \cite{hou2024comprehensive,liu2024kan,schmidt2021kolmogorov}.

Liu et al. \cite{liu2024kan} extended the Kolmogorov-Arnold Network (KAN) to networks of arbitrary depth, building on the Kolmogorov-Arnold representation theorem by unifying the outer functions \( \phi_q \) and inner functions \( \varphi_{q,p} \) into a series of KAN layers. In this extended form, the network architecture is defined by an integer array \([n_0, n_1, \dots, n_L]\), where \( n_L \) represents the number of neurons in the \( L \)-th layer. Each layer in the KAN is structured to transform an input vector \( x_l \) of \( n_l \) dimensions into an output vector \( x_{l+1} \) of \( n_{l+1} \) dimensions.

\begin{equation}
x_{l+1} = 
\begin{pmatrix}
\varphi_{1,1,l} & \cdots & \varphi_{1,n_l,l} \\
\vdots & \ddots & \vdots \\
\varphi_{n_{l+1},1,l} & \cdots & \varphi_{n_{l+1},n_l,l}
\end{pmatrix}
x_l
\label{eq:xl_output}
\end{equation}

Here, the function matrix for each layer \( l \) is denoted by \( \Phi_l \), where

\begin{equation}
\Phi_l = 
\begin{pmatrix}
\varphi_{l,1,1(\cdot)} & \cdots & \varphi_{l,1,n_l(\cdot)} \\
\vdots & \ddots & \vdots \\
\varphi_{l,n_{l+1},1(\cdot)} & \cdots & \varphi_{l,n_{l+1},n_l(\cdot)}
\end{pmatrix}.
\label{eq:phi_matrix}
\end{equation}

Within each \( l \)-th layer, the activation value of a neuron in the next layer is computed as the sum of all post-activation values from the previous layer. This hierarchical composition enables KAN to approximate complex multivariate functions through a series of recursive transformations. Mathematically, this process can be described as:

\begin{equation}
f(x) = \sum_{i_L=1}^{n_L} \phi_{i_L} \left( x_{i_L}^{(L)} \right),
\end{equation}

where each intermediate representation \( x_{i_L}^{(L)} \) is recursively defined by summing over non-linear transformations applied to outputs from the preceding layer:

\begin{equation}
x^{(L)}_{i_L} = \sum_{i_{L-1}=1}^{n_{L-1}} \varphi_{L-1, i_L, i_{L-1}} \left( x^{(L-1)}_{i_{L-1}} \right).
\end{equation}

In Liu et al.'s implementation, each KAN layer consists of a combination of spline functions and SiLU activations, which enhances flexibility in function approximation. By leveraging a variety of basis functions, including Legendre and Chebyshev polynomials, as well as Gaussian radial distribution functions, KAN can efficiently capture complex relationships within the data while maintaining interpretability and computational efficiency.

A general KAN network comprised of \( L \) layers can be written as:

\begin{equation}
\text{KAN}(x) = \left( \Phi_{L-1} \circ \Phi_{L-2} \circ \Phi_{L-3} \circ \dots \circ \Phi_1 \circ \Phi_0 \right) x,
\label{eq:layer_output}
\end{equation}

where each layer represents a transformation from one dimensionality to another, reducing the complexity of functions by stacking KAN layers. The transformation at each layer uses a matrix of univariate functions, rather than traditional weight matrices. This is formally given as:

\begin{equation}
x_{l+1,j} = \sum_{i=1}^{n_l} \phi_{l,j,i}(x_{l,i}),
\label{eq:layer_output_2}
\end{equation}

where \( \phi_{l,j,i} \) are spline-based univariate activation functions. The final output is generated through these successive transformations across layers \cite{liu2024kan}.

Figure \ref{fig:schematic_of_kan} provides a schematic representation of the Kolmogorov-Arnold Network (KAN) architecture, encapsulating the layered transformation process discussed above. Each layer in the KAN framework applies learnable, spline-based functions to inputs, replacing traditional fixed weights with adaptable non-linear mappings; enabling KANs to approximate complex, non-linear functions by capturing intricate dependencies across multiple layers. As a result, the KAN structure enhances symbolic function discovery capabilities while simultaneously improving computational efficiency, making it suitable for high-dimensional and complex data modeling tasks.

\underline{Approximation of Complex Functions in High-Dimensional Spaces:}

KAN leverages this theorem by approximating the univariate functions \( \varphi_{q,p}(x_p) \) and outer functions \( \phi_q \) using learnable splines. This allows KAN to dynamically adapt to the data patterns, as opposed to traditional neural networks like MLPs, which have fixed activation functions. The learnable nature of the splines allows KAN to approximate even complex, non-smooth functions that are otherwise challenging to capture with fixed activation networks \cite{hou2024comprehensive, liu2024kan, schmidt2021kolmogorov}.

KAN operates by replacing the weight matrices typically found in MLPs with these learnable univariate functions, transforming the output in a flexible manner. The univariate functions are parameterized as splines, which are piecewise polynomial functions that can adapt locally without losing global smoothness \cite{kilani2024kolmogorov}.

\underline{Interpretability and Scaling:}

KANs offer improved interpretability over traditional MLPs by making the functional mappings more transparent. Each univariate function is represented as a B-spline, allowing for fine control and clear visualization of how individual variables contribute to the final function. This interpretability makes KANs well-suited for scientific discovery tasks, where the goal is not only to approximate a function but also to gain insights into its structure.

Additionally, KANs leverage neural scaling laws that allow them to generalize well with fewer parameters compared to traditional deep learning models. The spline-based structure of KANs enables them to capture complex, high-dimensional relationships while avoiding the curse of dimensionality, which typically plagues models relying solely on fixed activation functions \cite{liu2024kan}.

\vspace{0.5cm} 

\underline{KAN Extension:}

KAN's flexibility and interpretability have led to a range of adaptations suited to specialized applications, from functional basis extensions to temporal and graph-based analyses. These KAN variants enhance the model's accuracy, computational efficiency, and capacity for capturing complex relationships. Extensions include versions that leverage orthogonal polynomials, rational function bases, and wavelet transformations for improved mathematical representation; time series adaptations that dynamically respond to temporal patterns; and graph-based structures tailored for graph-structured data processing. Together, these KAN adaptations expand the model’s versatility, providing targeted improvements across diverse domains.

\begin{figure}[h]
  \centering
  \includegraphics[width=\linewidth]{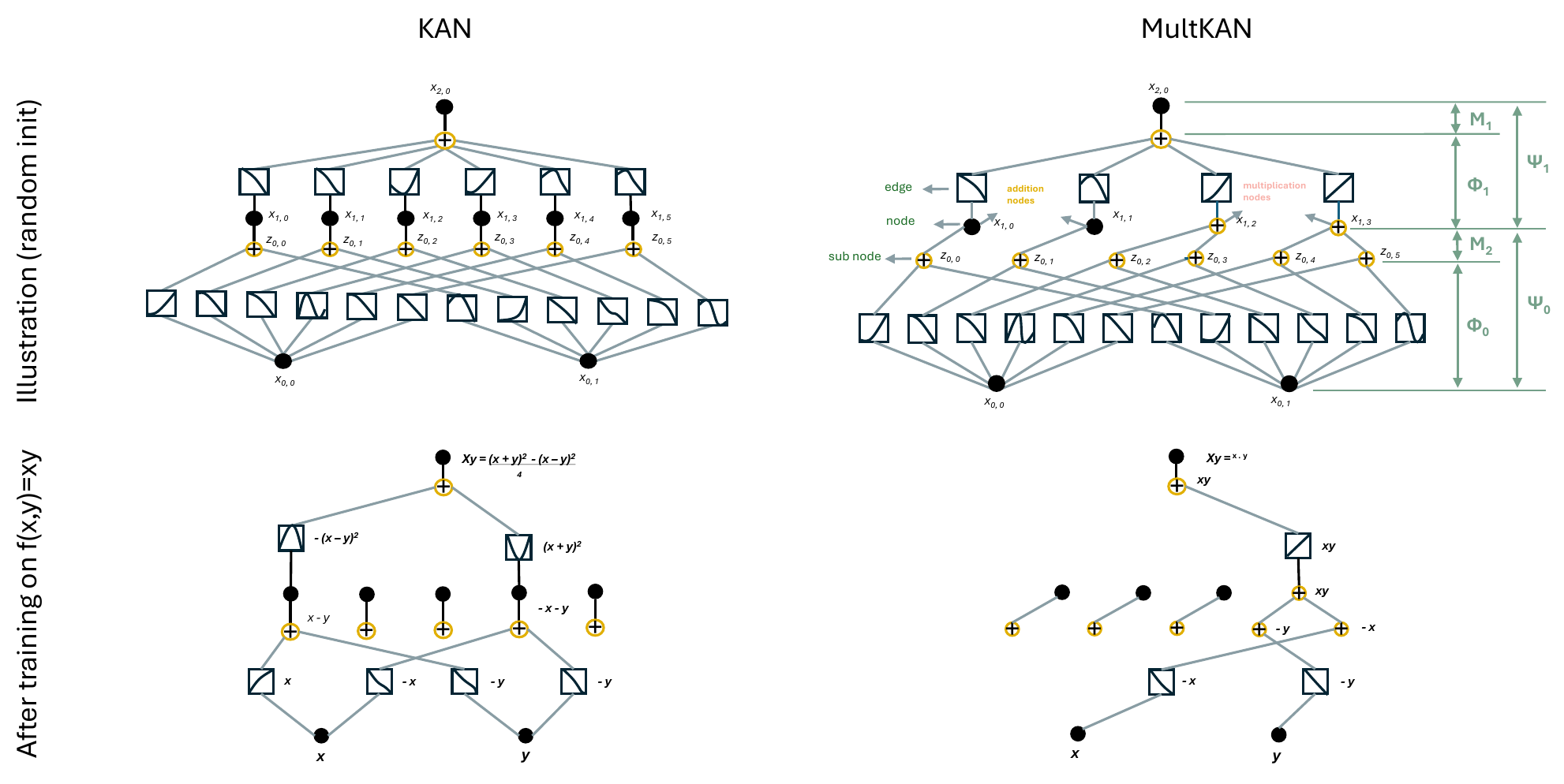}
  \caption{Comparing KAN and MultKAN diagrams \cite{liu2024kan}}
  \label{fig:kan_vs_multkan}
\end{figure}

Figure \ref{fig:kan_vs_multkan} illustrates a structural comparison between the KAN and the enhanced MultKAN architecture. The top section displays the layout of each network, highlighting how MultKAN introduces additional multiplication layers, which allow it to capture multiplicative relationships more efficiently. The bottom section demonstrates the training outcomes on the function \( f(x, y) = xy \), where KAN utilizes two addition nodes to approximate multiplication, while MultKAN achieves the same task with a single multiplication node. This adaptation enables MultKAN to directly represent multiplicative structures, thereby enhancing computational efficiency and interpretability in symbolic function discovery.

The Partial Differential Equation Kolmogorov-Arnold Network (PDE-KAN) enhances the original KAN by incorporating physics-informed elements suited for solving differential equations. Unlike traditional KANs, which are limited to general function approximation, PDE-KANs employ physics-based constraints and loss functions, enhancing interpretability and accuracy in modeling physical phenomena described by PDEs. This adaptation enables PDE-KANs to tackle both forward and inverse problems in computational physics, making them more effective for high-complexity tasks involving boundary and initial conditions. By leveraging PDE forms such as the energy, strong, and inverse forms, PDE-KANs achieve improved convergence rates and solution accuracy, demonstrating significant advantages over conventional neural network approaches like MLPs \cite{shukla2024comprehensive,wang2024kolmogorov}.

While PDE-KAN is tailored for solving differential equations by incorporating physics-informed loss functions that enforce boundary and domain constraints \cite{wang2024kolmogorov}, Rational KAN (rKAN), proposed by Aghaei \cite{aghaei2024rkan}, extends KAN by using rational function bases—specifically employing Padé approximations and rational Jacobi functions—to significantly improve performance in regression and classification tasks. This method enhances model accuracy and computational efficiency through optimized activation functions and more efficient parameter update mechanisms. In rKAN, the use of rational functions—via polynomial divisions and mapped Jacobi functions—enables it to capture sharp peaks and rapid changes in data more effectively than traditional KAN. The rational Jacobi approach also extends function approximation over semi-infinite or infinite domains, improving rKAN’s versatility for tasks requiring high precision across broad input spaces, such as physics-informed problems with complex boundary conditions. The overall rKAN formulation is expressed as: 

\begin{equation}
F(\xi) = \sum_{q=1}^{2n+1} \Phi_q \left( \sum_{p=1}^n \varphi_{q,p}(\xi_p) \right)
\label{eq:multkan_formula}
\end{equation}

where \( \varphi_{q,p}(\cdot) \) are rational functions based on Padé or Jacobi mappings. Here, \( \Phi_q(\cdot) \) serves as an outer activation or aggregation function, providing flexibility in capturing complex interactions across input dimensions. This layered structure allows rKAN to generalize effectively across various types of data and tasks, offering significant improvements in precision and interpretability.

The Wav-KAN adapts the KAN architecture by incorporating wavelet functions as learnable activation functions, enabling nonlinear mapping of input spectral features and effectively capturing multi-scale spatial-spectral patterns through dilation and translation. This approach allows Wav-KAN to isolate significant patterns at various scales, enhancing its ability to filter out noise while retaining critical features, which is particularly useful in hyperspectral image classification. Using the Continuous Wavelet Transform and Discrete Wavelet Transform, Wav-KAN captures both high-frequency and low-frequency components, improving interpretability and robustness compared to traditional models. As demonstrated by Seydi et al. \cite{seydi2024unveiling}, Wav-KAN significantly outperformed traditional MLP and Spline-KAN models, achieving notable improvements in classification accuracy on benchmark datasets such as Salinas, Pavia, and Indian Pines. This model also enhances computational efficiency by reducing the number of necessary parameters without sacrificing precision, making Wav-KAN an efficient and powerful solution for handling the high-dimensional, correlated nature of hyperspectral data.

Temporal KAN (T-KAN) and Multi-Task KAN (MT-KAN) are specialized KAN variants developed for time series applications \cite{xu2024kolmogorov}. T-KAN, designed for univariate time series data, utilizes learnable univariate activation functions that dynamically adapt to nonlinear relationships and capture complex temporal patterns, allowing it to effectively handle variations across time. The architecture models relationships between consecutive time steps, predicting future values while tracking concept drift, a capability particularly valuable in financial forecasting and energy demand prediction. The T-KAN output at time \( t+T \), denoted \( S_{t+T} \), is given by:

\begin{equation}
S_{t+T} = \sum_{q=1}^{2n+1} \Phi_q \left( \sum_{p=1}^n \varphi_{q,p} \left( S_{t-h+p} \right) \right)
\label{eq:TKAN_output}
\end{equation}
where \( S_{t-h+p} \) represents past observations, and \( \varphi_{q,p} \) are spline-parametrized functions learned during training to model nonlinear temporal dependencies. Additionally, symbolic regression enhances T-KAN’s interpretability by generating human-readable expressions that reveal the underlying dependencies between time steps.

Building on T-KAN’s capabilities, MT-KAN extends this approach to handle multivariate time series by introducing a shared network structure that enables multi-task learning across related tasks. This model effectively captures inter-variable relationships, enhancing predictive accuracy in applications like electricity load forecasting and air quality monitoring, where multivariate dependencies are essential. MT-KAN improves performance by leveraging inter-task relationships, optimizing feature representations across tasks to boost prediction accuracy with reduced training data requirements. This makes MT-KAN an efficient choice for complex, interdependent time series data.

\begin{figure}[h]
  \centering
  \includegraphics[width=\linewidth]{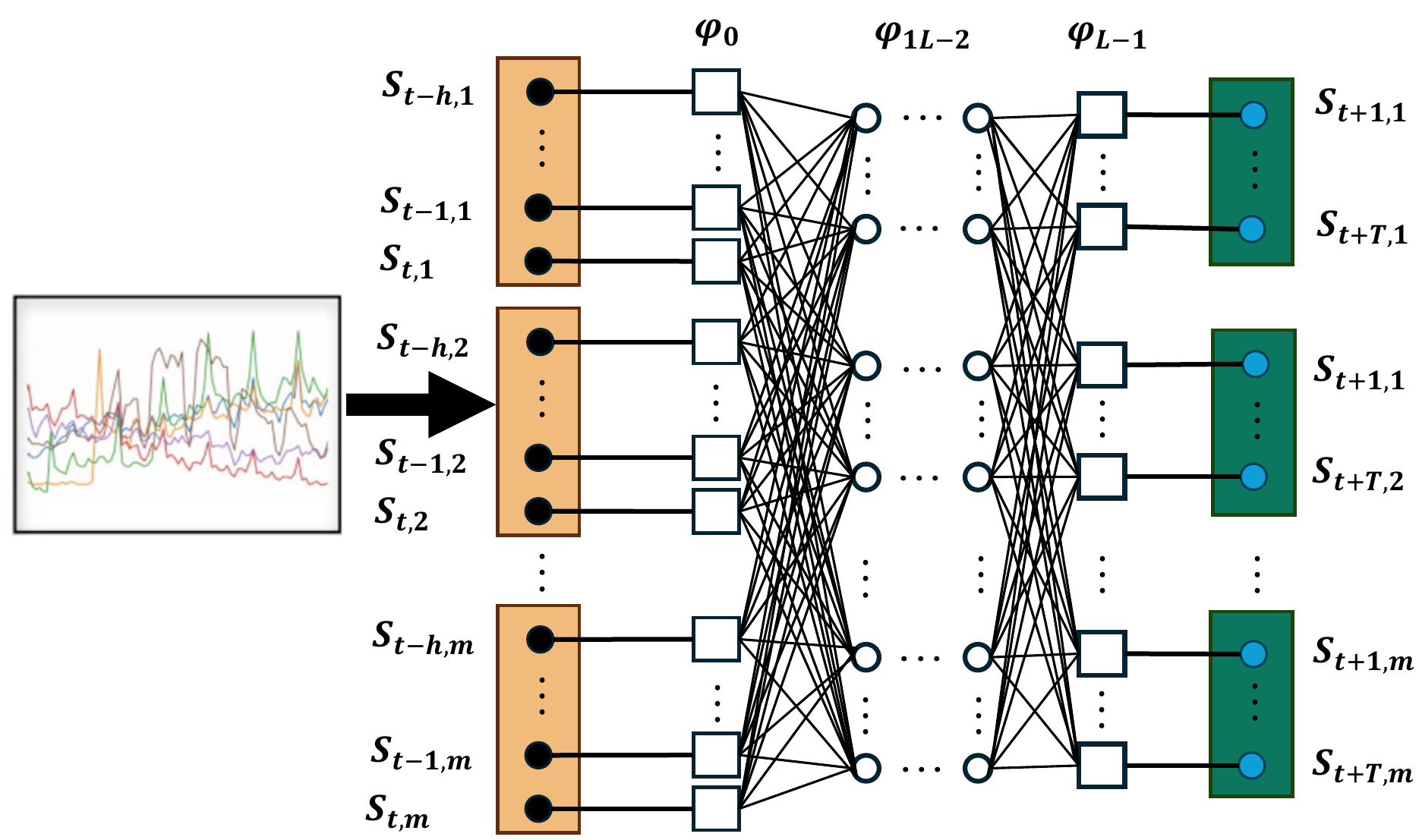}
  \caption{MT-KAN architecture for multivariate time series  \cite{xu2024kolmogorov}}
  \label{fig:mt_kan_achitecture}
\end{figure}

Figure \ref{fig:mt_kan_achitecture} shows the MT-KAN architecture for multivariate time series, where past values of multiple variables are processed through shared network layers \( \Phi_0, \Phi_{1:L-2}, \Phi_{L-1} \) to capture temporal and cross-variable dependencies. This setup allows MT-KAN to model complex interactions between variables and improve forecasting accuracy for interdependent tasks. The final layer outputs future values for each variable over the predicted horizon.

SigKAN enhances KAN's capabilities in time series prediction by incorporating path signatures, which capture essential geometric features of time series paths \cite{inzirillo2024sigkan}. By integrating these signatures with KAN’s output, SigKAN provides a more comprehensive representation of temporal data, effectively capturing complex time series patterns through iterated integrals. The core architecture of SigKAN includes a Learnable Path Signature Layer that computes path signatures for each input sequence, allowing the network to capture sequential dependencies and intricate path structures. These signatures are combined with the KAN output through a Gated Residual KAN Layer (GRKAN), which modulates information flow by applying weighting mechanisms to enhance relevant features and suppress noise. The output of SigKAN can be described by the following equation:
\begin{equation}
y = \psi \odot \text{KAN}(X),
\label{eq:sigkan_output}
\end{equation}
where \( \psi = \text{SoftMax}(\text{GRKAN}(S(X))) \) is the weighted output from the Gated Residual KAN layer, and \( S(X) \) represents the path signature of the input sequence \( X \). The path signature \( S(X) \) includes iterated integrals of the input sequence, capturing complex temporal patterns across different scales.

\begin{figure}[h]
  \centering
  \includegraphics[width=0.5\linewidth]{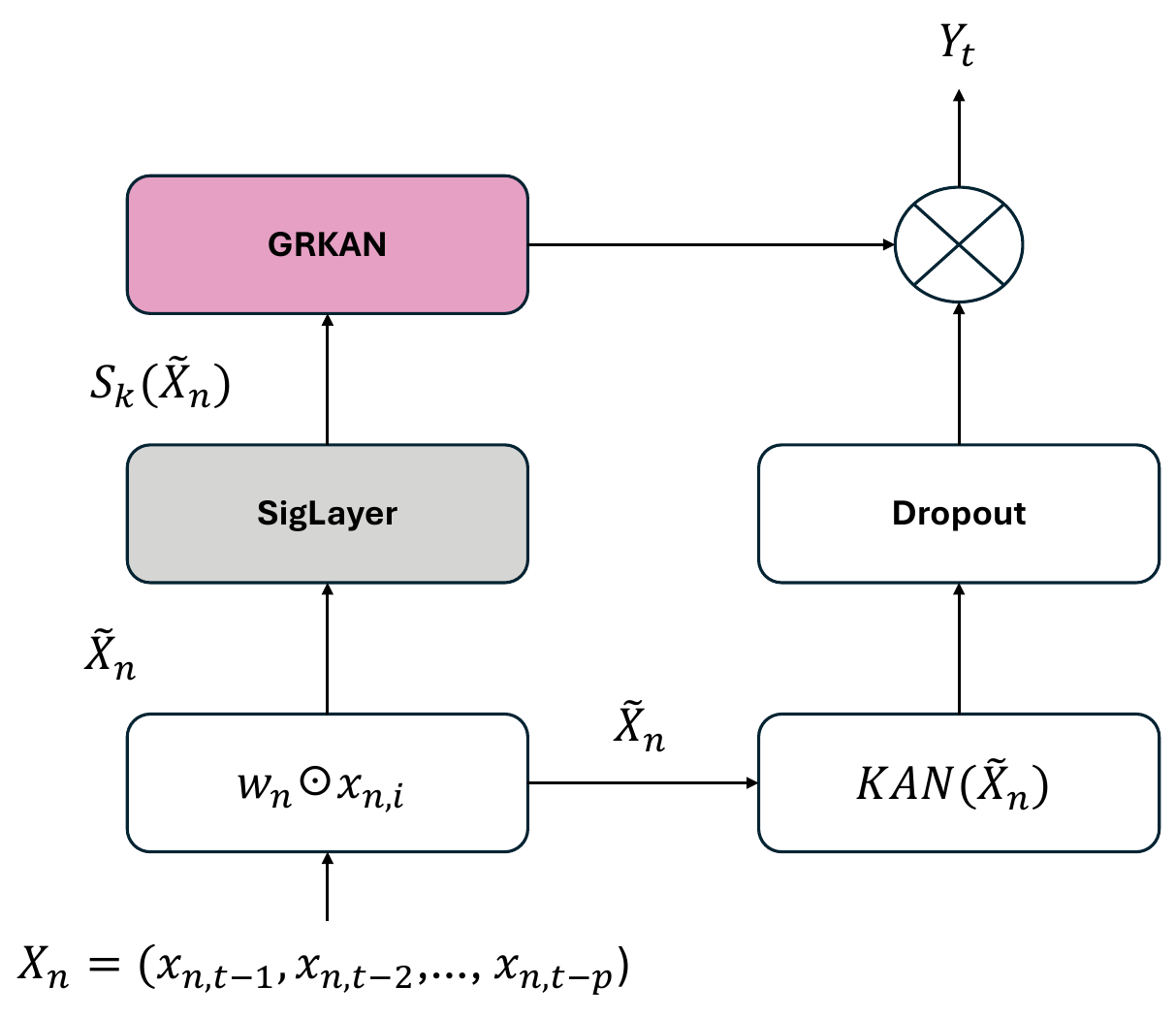}
  \caption{SigKAN architecture \cite{inzirillo2024sigkan}}
  \label{fig:sigkan}
\end{figure}

Figure \ref{fig:sigkan} illustrates the SigKAN architecture, where input data \( X_n = (x_{n,t-1}, x_{n,t-2}, \dots, x_{n,t-p}) \) is first processed by learnable weight coefficients \( w_n \) to create \( \hat{X}_n \). This representation passes through the SigLayer to compute path signatures \( S_k(\hat{X}_n) \), capturing essential path features. The signatures are then fed into the GRKAN module, which weights and modulates this information. Simultaneously, \( \hat{X}_n \) is passed to the KAN layer, and a Dropout layer is applied for regularization. The final output \( Y_t \) is obtained by combining the GRKAN-weighted signatures with the KAN output, offering a robust prediction that leverages both path signature geometry and KAN's functional approximation capabilities.

This integration of path signatures with KAN’s functional approximation allows SigKAN to adapt dynamically to nonlinear relationships in time series, making it effective in tasks like financial modeling and multivariate forecasting. As demonstrated by Inzirillo and Genet \cite{inzirillo2024sigkan}, SigKAN significantly outperforms traditional KAN models, achieving both improved accuracy and a deeper understanding of temporal dependencies.

Graph KAN (GKAN) extends the KAN framework to graph-structured data by introducing learnable univariate functions on graph edges, replacing the fixed convolutional structure of traditional Graph Convolutional Networks (GCNs) \cite{kiamari2024gkan}. GCNs operate by iteratively aggregating and transforming feature information from local neighborhoods within a graph, effectively capturing both node features and graph topology. This approach, pioneered by \cite{kipf2016semi}, has proven effective for various applications, including node classification and recommendation systems. However, GCNs rely on fixed convolutional filters, which limits their flexibility in handling complex, heterogeneous graphs.

\begin{figure}[h]
  \centering
  \includegraphics[width=\linewidth]{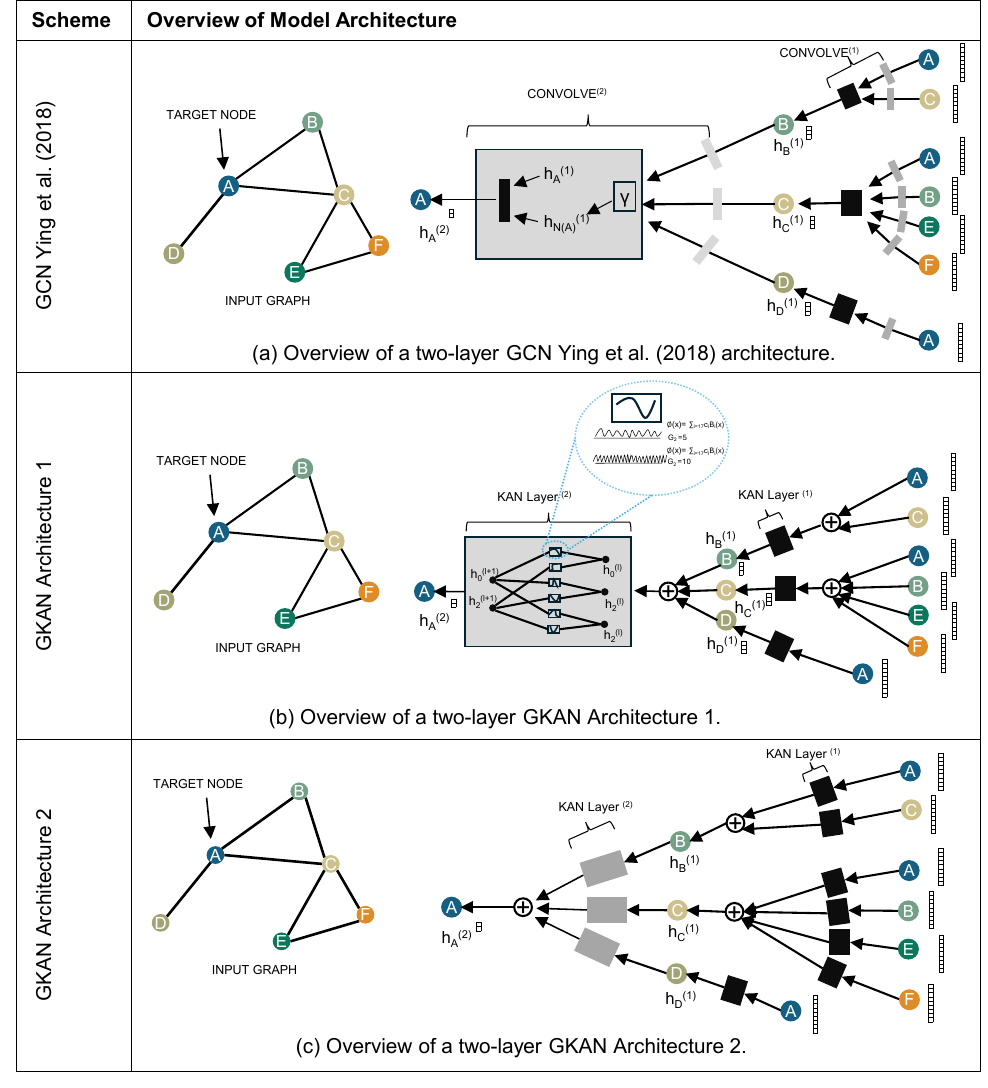}
  \caption{Comparison of different GKAN model architectures  \cite{kiamari2024gkan}}
  \label{fig:gkan_achitecture}
\end{figure}

To address this limitation, GKAN introduces two primary architectures: Architecture 1, which aggregates node features before applying KAN layers, allowing learnable activation functions to capture complex local relationships, and Architecture 2, which places KAN layers between node embeddings at each layer before aggregation, allowing for dynamic adaptation to changes in graph structure. Formally, in Architecture 1, the embedding of nodes at layer \( \ell+1 \) is represented as:
\begin{equation}
H^{(\ell+1)}_{\text{Archit.1}} = \text{KANLayer}( \hat{A} H^{(\ell)}_{\text{Archit.1}} )
\end{equation}
where \( \hat{A} \) is the normalized adjacency matrix, and \( H^{(0)}_{\text{Archit.1}} = X \) (input features). In Architecture 2, the process is reversed:
\begin{equation}
H^{(\ell+1)}_{\text{Archit.2}} = \hat{A} \, \text{KANLayer}( H^{(\ell)}_{\text{Archit.2}} ).
\end{equation}
This flexible setup enables GKAN to adapt effectively to large-scale and heterogeneous graph data by optimizing feature representation across evolving graph structures.

Figure \ref{fig:gkan_achitecture} provides a visual comparison between the traditional GCN and the two proposed GKAN architectures. Part (a) illustrates the GCN setup, where convolutional layers are applied directly on node embeddings. Part (b) demonstrates GKAN Architecture 1, which aggregates node features before passing them through KAN layers, while Part (c) depicts Architecture 2, where KAN layers are applied to individual node embeddings prior to aggregation. This figure highlights how each GKAN architecture processes graph data differently, enabling greater adaptability and flexibility compared to standard GCNs.

\subsection{Optimization Strategies}

Optimization in KAN is essential due to the complexity of the high-dimensional function approximations they perform. KAN’s power lies in its ability to decompose multivariate functions into univariate spline functions, but the effectiveness of this process depends on the optimization of these splines. Optimization adjusts the control points and knots of the splines to minimize errors between predicted and actual outputs, allowing the model to capture intricate data patterns. However, the high-dimensionality and complexity of these approximations introduce several challenges:

\begin{itemize}
    \item \textbf{Non-linear Parameter Space:} Unlike traditional neural networks, KAN involves adjusting spline parameters, which makes the optimization landscape non-linear and harder to navigate.
    \item \textbf{Curse of Dimensionality:} As the number of dimensions grows, the number of parameters increases, leading to potential overfitting and slower convergence.
    \item \textbf{Computational Overhead:} The flexibility of learnable splines increases the computational burden during training, making optimization slower and more resource-intensive \cite{hou2024comprehensive}.
\end{itemize}

Given the challenges of optimizing KAN, several techniques have been employed to improve convergence and performance:

\begin{itemize}
    \item \textbf{Gradient Descent and its Variants:} Since KAN parameters (spline control points) are optimized using gradient-based techniques, variants of gradient descent like Stochastic Gradient Descent (SGD) with momentum are often used to help smooth the optimization path and avoid local minima. These techniques mitigate the difficulties of non-linear optimization by using momentum to escape saddle points \cite{hou2024comprehensive}.
    
    \item \textbf{Adam Optimizer:} Adam is another popular optimization technique that combines momentum and adaptive learning rates, making it highly effective for training KAN models. Adam’s ability to adjust the learning rate for each parameter individually is beneficial for optimizing KAN’s complex spline functions \cite{hou2024comprehensive}.
    
    \item \textbf{Regularization Techniques:} To prevent overfitting in high-dimensional spaces, L2 regularization and dropout are commonly used. These techniques help constrain the flexibility of the splines and prevent them from fitting the noise in the training data \cite{hou2024comprehensive}.
    
    \item \textbf{Batch Normalization:} To stabilize training and speed up convergence, batch normalization is often applied in KAN layers. This technique helps address the vanishing or exploding gradient problem, which is common in deep and complex networks \cite{hou2024comprehensive}.
\end{itemize}

Convergence during KAN training is a known challenge due to the high dimensionality and non-linear optimization landscape. Several papers have highlighted specific issues related to sensitivity to initialization and slow convergence in high-dimensional spaces:

\begin{itemize}
    \item \textbf{Sensitivity to Initialization:} Poor initialization of spline parameters can lead to suboptimal convergence or cause the optimization to get stuck in local minima. Research has shown that careful initialization strategies, such as He initialization or Xavier initialization, can mitigate these issues by providing a better starting point for optimization \cite{hou2024comprehensive, liu2024kan}.
    
    \item \textbf{Slow Convergence:} Due to the large number of learnable parameters in KAN (especially when dealing with high-dimensional data), optimization can converge very slowly. This is compounded by the need to optimize both the shape and position of the splines, making the process more complex than standard neural networks. Advanced techniques such as second-order optimizers (e.g., L-BFGS) have been proposed to speed up convergence by leveraging curvature information of the loss landscape \cite{hou2024comprehensive, liu2024kan}.
    
    \item \textbf{Regularization and Dropout:} Overfitting is a common issue due to the flexible nature of splines. To address this, researchers have proposed dropout during training, which helps improve generalization by randomly removing units and their connections, reducing overfitting risks \cite{hou2024comprehensive}.
    
    \item \textbf{Optimization Instabilities:} Researchers have also discussed optimization instabilities in KAN, especially in high-dimensional spaces. One common issue is the tendency of optimization algorithms to converge to local minima rather than global ones, particularly when dealing with spline-based architectures. Early stopping and learning rate schedules have been recommended to help address these convergence difficulties by preventing overfitting and ensuring smoother optimization \cite{hou2024comprehensive, liu2024kan}.
\end{itemize}

KANs incorporate unique optimization strategies due to their spline-based architecture and dynamic activation functions. The key optimization approaches are listed below:

\begin{enumerate}
    \item \textbf{Spline-based Learnable Weights:} The weights in KANs are not fixed linear transformations as in traditional models but are instead learnable spline functions. These weights are optimized using gradient-based methods like backpropagation, similar to other neural networks. However, specific adjustments are required to handle the B-spline functions used in KANs. To address this:
    \begin{itemize}
        \item Spline grid updates are implemented during training to adjust the locations of knots in B-splines, allowing the splines to adapt dynamically as the network learns.
        \item Gradient descent is employed to optimize the coefficients of the spline functions, with methods such as SGD or Adam being commonly used.
    \end{itemize}
    
    \item \textbf{Variance-Preserving Initialization:} KAN models require careful initialization of the spline coefficients to ensure stable training. A variance-preserving initialization is often used to maintain the variance of activations across layers. This ensures that the model does not suffer from vanishing or exploding gradients during optimization, which is crucial given the non-linear nature of KANs.
    
    \item \textbf{Residual Activation Functions:} To improve convergence during training, KANs often employ residual activations. These are combinations of basic activation functions, \( b(x) \) (e.g., SiLU) and learned spline functions. This structure is inspired by the residual connections used in CNNs, which have proven effective in accelerating convergence and mitigating vanishing gradient issues:
\begin{equation}
\phi(x) = w_b \cdot b(x) + w_s \cdot \text{spline}(x)
\label{eq:activation_function}
\end{equation}

    The weights \( w_b \) and \( w_s \) control the contributions of the basic activation and the spline, respectively. This structure introduces non-linearity via the basic activation, while the spline models more complex, fine-grained relationships in the data. Together, they improve convergence, allowing KANs to learn smoothly and effectively balance both components during training.
    
    \item \textbf{Efficient Splines for Faster Computation:} Given that traditional B-splines can be computationally expensive, some variants of KAN, like FastKAN, replace B-splines with more efficient Gaussian RBFs. This reduces the computational complexity and speeds up both forward and backward passes during optimization.
\end{enumerate}

Table \ref{tab:optimization} compares the key optimization features of CNNs, RNNs, and KANs. CNNs, introduced by Yann LeCun in 1989 \cite{lecun1989backpropagation}, employ fixed linear weights and non-linear activation functions (such as ReLU), making them highly effective for image recognition tasks like classification and object detection. RNNs, conceptualized by John Hopfield in 1982 \cite{hopfield1982neural} and later refined by Ronald J. Williams and David E. Rumelhart \cite{rumelhart1986learning}, are designed for sequential data processing, using fixed linear weights and non-linear gated mechanisms to model temporal dependencies in tasks such as language modeling and time series forecasting. In contrast, KANs extend neural network capabilities by incorporating learnable spline-based weights and activations, providing greater flexibility and interpretability compared to CNNs and RNNs. 

KANs also utilize variance-preserving initialization and offer computational efficiency, especially when optimized with methods like FastKAN. Meanwhile, CNNs use random initialization, and RNNs often rely on orthogonal or identity initialization to maintain stability, with the latter being more computationally intensive due to backpropagation through time. Additionally, KANs leverage residual activations through spline bases, while residual connections are common in CNNs (e.g., ResNets) and less typical in RNNs, though they appear more frequently in models like LSTMs.

\thispagestyle{empty}
\begingroup
\fontsize{8pt}{8pt}\selectfont
\begin{table}[h!]
  \centering
  \caption{Optimization Strategies in KAN vs Other Models}
  \label{tab:optimization}
  \begin{tabular}{p{3cm} p{4cm} p{3cm} p{3cm}}
    \toprule
    \textbf{Optimization Feature} & \textbf{KAN} & \textbf{CNN} & \textbf{RNN} \\
    \midrule
    Weight Representation & Learnable spline-based weights & Fixed linear weights & Fixed linear weights \\
    \midrule
    Activation Functions & Spline-based activation on edges & Non-linear activations on nodes (e.g., ReLU) & Non-linear activations with gated mechanisms \\
    \midrule
    Initialization & Variance-preserving initialization of spline grids & Random initialization & Orthogonal or identity initialization for stability \\
    \midrule
    Residual Connections & Residual activation with spline-basis & Common (e.g., ResNets) & Less common; more typical in LSTMs \\
    \midrule
    Computational Complexity & Efficient with spline optimizations (e.g., FastKAN) & Moderate (requires large networks for performance) & High (requires backpropagation through time) \\
    \bottomrule
  \end{tabular}
\end{table}
\endgroup

\subsection{Regularization Techniques}

Kolmogorov–Arnold Networks (KANs) utilize several regularization techniques to mitigate overfitting, a critical challenge in machine learning models. The most common techniques include the following:

\begin{enumerate}
    \item \textbf{Dropout:}
    Dropout is a popular regularization method where randomly selected neurons are ignored during training. This prevents the network from becoming too reliant on specific neurons, which can lead to overfitting. By dropping out a random fraction of neurons, the network is forced to learn more general features that are robust to different inputs. Although dropout is common in traditional neural networks, its implementation in KANs depends on their architecture and how the inner and outer functions are modeled.
\begin{equation}
h_i = \text{Dropout}(f_i(x), p)
\label{eq:dropout} 
\end{equation}

    where \( h_i \) is the hidden layer output after dropout, \( f_i(x) \) is the function applied by the hidden layer, and \( p \) is the dropout probability.
    
    \item \textbf{Weight Regularization (L2 Norm or Ridge Regularization):}
    In KANs, weight regularization is applied to penalize large weights, which could indicate overfitting. The L2 norm, also known as Ridge regularization, adds a penalty to the loss function proportional to the square of the magnitude of the weights. This technique encourages the network to maintain smaller weights, resulting in a smoother model that is less likely to overfit the training data.
\begin{equation}
\text{Loss} = \text{MSE} + \lambda \sum w_i^2
\label{eq:loss_function} 
\end{equation}

    where \( \lambda \) is the regularization strength, and \( w_i \) are the network weights.
    
    \item \textbf{Early Stopping:}
    Early stopping is a regularization technique that monitors the model's performance on the validation dataset during training. If the model's performance on the validation set starts to degrade while continuing to improve on the training set, training is stopped. This prevents the model from becoming overly fitted to the training data. KAN models can implement this technique, as shown in various deep learning tasks \cite{aghaei2024rkan}.
    
    \item \textbf{Constraints on Inner and Outer Functions:}
    In KANs, constraints are often applied to the inner and outer functions to ensure they maintain certain properties that improve generalization. For example, the smoothness or continuity of the functions is often controlled through constraints on their derivatives. These constraints help the model avoid fitting noise and ensure that the functions remain interpretable and generalizable \cite{guilhoto2024deep}.
    
    \item \textbf{Gradient Clipping:}
    Gradient clipping prevents the problem of exploding gradients during backpropagation by limiting the size of gradients during training. This ensures that the training process remains stable and prevents the model from converging to sharp, overfitting-prone solutions. Though not unique to KANs, gradient clipping is another useful regularization technique in complex architectures like these \cite{guilhoto2024deep}.
    
    \item \textbf{Sparsity Constraints:}
    KANs can also incorporate sparsity regularization, where additional terms are added to the loss function to encourage sparse activations. This means that only a few neurons are active at a time, reducing the model's complexity and helping prevent overfitting by making the model more interpretable and efficient \cite{alter2024robustness}.
\end{enumerate}

\section{Applications of KAN}
\subsection{Field-Specific Applications}
KANs have emerged as a transformative neural network architecture with versatile applications across various domains, including time series forecasting, computational biomedicine, graph learning, survival analysis, power systems, and physics-informed neural networks for scientific computing. Inspired by the Kolmogorov-Arnold representation theorem, KANs replace traditional linear weights with adaptive activation functions, typically spline-parametrized univariate functions, that allow them to dynamically learn and approximate complex, high-dimensional relationships more efficiently than conventional models. This unique approach has demonstrated significant performance advantages in fields that require high-dimensional function approximation, enhancing both predictive accuracy and model interpretability.

In time series analysis, KANs have proven particularly effective in satellite traffic forecasting, where their adaptive activation functions capture intricate temporal patterns and outperform traditional models with fewer parameters \cite{vaca2024kolmogorov}. Similarly, in computational biomedicine, the integration of structural knowledge into KANs enhances model reliability, reduces training data requirements, and mitigates the risk of generating spurious predictions \cite{samadi2024smooth}. TKAN extend this capacity by combining KAN’s architecture with memory mechanisms, making them ideal for complex sequential data management in domains such as finance and healthcare, where long-term dependencies are critical \cite{genet2024tkan}.

\begin{figure}[htp]
    \centering
    \includegraphics[width=0.98\linewidth]{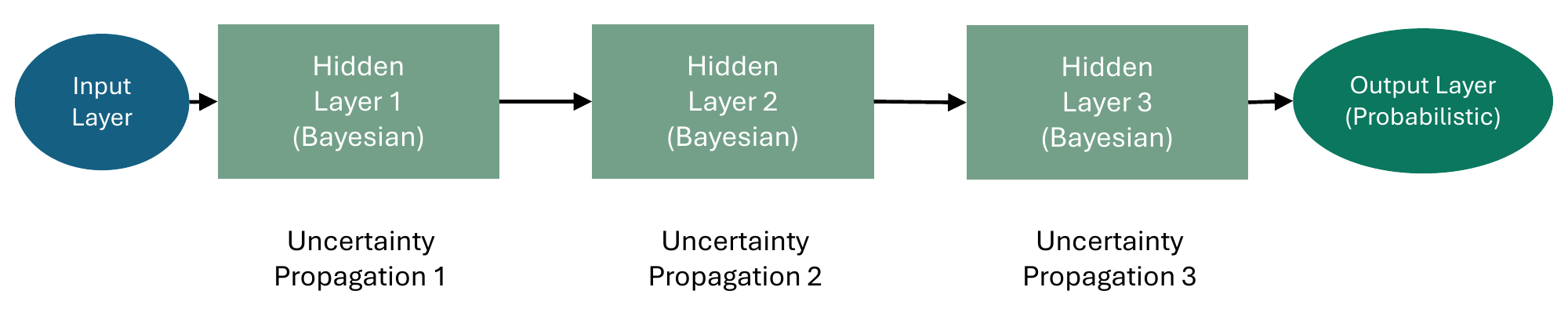}
    \caption{Hierarchical Bayesian structure in BKANs, illustrating uncertainty propagation through each layer to produce probabilistic outputs for robust decision-making \cite{hassan2024bayesian}}
    \label{fig:BKAN}
\end{figure}

Expanding KAN's reach into healthcare, Bayesian KANs (BKANs) offer interpretable, uncertainty-aware predictions essential for medical diagnostics. BKANs enhance decision support by representing both aleatoric and epistemic uncertainty in predictions, which is critical in fields handling imbalanced or limited data such as medical diagnosis \cite{hassan2024bayesian}. Figure \ref{fig:BKAN} shows the Bayesian hierarchical structure of BKANs, where uncertainty is propagated through each Bayesian layer to yield probabilistic outputs, supporting robust decision-making in medical diagnostics. In survival analysis, CoxKAN, a KAN-based framework specifically tailored for Cox proportional hazards modeling, facilitates high-performance survival analysis with automatic feature selection and symbolic formula extraction, enabling effective biomarker identification and complex variable interaction discovery in healthcare datasets \cite{knottenbelt2024coxkan}.

KAN’s flexibility also enables its reformulation as RBF networks, a modification known as FastKAN, which accelerates approximation processes in fields like pattern recognition by leveraging Gaussian kernels \cite{li2024kolmogorov}. The architecture’s modular nature further allows it to scale efficiently, as demonstrated in tasks involving graph and tabular data where it outperforms MLPs in interpretability and parameter efficiency \cite{zhang2024generalization}. In multivariate time series forecasting, variants like T-KAN and MT-KAN enable adaptive concept drift detection and improved forecasting in dynamic environments, showcasing KAN’s capacity to reveal complex, nonlinear relationships in evolving data \cite{xu2024kolmogorov}. 

Applications in graph learning also benefit from KAN’s innovative design; for instance, in GNNs, KANs enhance node representation learning and improve regression accuracy in domains such as social networks and molecular chemistry \cite{bresson2024kagnns}. Despite their strengths, KANs are sensitive to noise, as shown in signal processing studies where noise-reduction techniques, like kernel filtering, have been used to retain model accuracy under noisy conditions, further extending KAN’s applications in fields where data irregularities are prevalent \cite{shen2024reduced}. 

In scientific computing, physics-informed KANs (PIKANs) have been adapted to solve PDEs, offering an interpretable and efficient alternative to MLP-based PINNs. PIKANs’ adaptive grid-dependent structure excels in applications requiring precision, such as fluid dynamics and quantum mechanics, where dynamic basis functions enable these models to capture complex physical processes with superior accuracy and computational efficiency \cite{rigas2024adaptive, shuai2024physics}. This diverse array of applications underscores KAN’s capacity to bridge computational demands and interpretability needs across fields, advancing the development of efficient, accurate, and transparent neural networks suited to domain-specific challenges.

\subsection{Performance Comparison}
KANs have emerged as a promising alternative to traditional neural network architectures like MLPs, CNNs, and RNNs, driven by their unique spline-based, learnable activation functions and the Kolmogorov-Arnold representation theorem. KANs replace conventional weight structures with univariate spline functions that enhance interpretability and accuracy, especially in applications requiring complex function approximation and adaptive prediction. Studies indicate that KANs often outperform MLPs in predictive accuracy and computational efficiency across various tasks, including electrohydrodynamic pump modeling, time series forecasting, and graph learning \cite{peng2024predictive, samadi2024smooth}.

In the predictive modeling of electrohydrodynamic (EHD) pumps, KANs demonstrated superior accuracy and interpretability over MLPs and Random Forests by more effectively capturing nonlinear relationships between parameters \cite{peng2024predictive}. Likewise, in time series forecasting, KAN-based models consistently outperformed traditional architectures with fewer parameters, making them highly efficient and suitable for real-time predictive tasks such as satellite traffic analysis \cite{vaca2024kolmogorov}. Further advancements in time series forecasting have been achieved by Temporal KANs (TKAN), which enhance long-term dependency handling by integrating LSTM layers, thereby addressing complex temporal patterns with greater accuracy than RNNs and MLPs, particularly in multi-step predictions \cite{genet2024tkan}. This architecture's efficiency is further supported by theoretical analyses showing lower generalization bounds and more favorable scaling laws compared to MLPs, particularly in low-dimensional tasks with compositional structures, reinforcing KANs’ role in computationally constrained settings \cite{hou2024comprehensive}.

However, KANs are not without limitations. While they excel in interpretability and accuracy, particularly for low-dimensional and structured data, their performance is notably sensitive to noise. In noisy environments, KANs may experience substantial performance degradation, necessitating complex noise-mitigation strategies like oversampling and kernel filtering, which can increase computational demands \cite{shen2024reduced}. In scientific applications, the KINN, a KAN variant, has proven particularly effective in solving multi-scale and heterogeneous problems in physics-informed deep learning. Here, KINN significantly outperforms MLPs in accuracy and convergence speed when solving PDEs, though it may struggle with complex boundary conditions where simpler MLPs show advantages \cite{wang2024kolmogorov}.

Despite these challenges, KANs have shown advantages in tasks such as graph regression and structured time-series forecasting, where interpretability and computational efficiency are prioritized \cite{bresson2024kagnns}. For image-related tasks, Convolutional KANs demonstrate similar levels of accuracy as CNNs but with far fewer parameters, making them highly resource-efficient without compromising performance on datasets like MNIST and Fashion-MNIST \cite{bodner2024convolutional}. Furthermore, the GKAN introduces KAN-based spline activations within GNNs, allowing for improved accuracy and inherent interpretability in graph tasks such as node classification and link prediction, outperforming traditional GNN models in interpretability-focused applications \cite{de2024kolmogorov}.

\section{Challenges and Limitations}
\subsection{Computational Complexity}
KANs have garnered increasing attention as a promising alternative to traditional neural network architectures, such as MLPs, due to their unique ability to decompose multivariate functions into simpler, univariate components. Despite these advantages, the implementation of KANs in high-dimensional spaces presents significant computational challenges, particularly in terms of optimization, training time, and resource usage. One of the primary issues is the non-convex nature of the optimization problem, which leads to slower convergence and longer training times compared to conventional neural networks. For instance, Bresson et al.\cite{bresson2024kagnns} noted that KANs struggle with high computational complexity in graph learning tasks, which increases training time when compared to MLPs. Additionally, KANs are particularly sensitive to noise in the training data, which further exacerbates computational inefficiencies. Shen et al. \cite{shen2024reduced} highlighted how the presence of noise significantly degrades KAN performance, necessitating costly noise-reduction techniques such as kernel filtering and oversampling, which further inflate the computational overhead. Similarly, Nagai and Okumura \cite{nagai2024kolmogorov} explored the use of KANs in molecular dynamics simulations and found that while KANs can reduce computational costs for low-dimensional tasks, their scalability is limited in high-dimensional settings due to the complexity of spline interpolation used for function approximation. 

Moreover, KANs tend to demand more computational resources than black-box models like MLPs. As Sun \cite{sun2024evaluating}  pointed out, KANs often require additional computational power and training time to handle complex, high-dimensional tasks, even though they offer advantages in interpretability and symbolic function generation. In addition, hardware implementation of KANs poses another significant challenge. A study by Tran et al. \cite{le2024exploring}  revealed that KANs consume substantially higher hardware resources than MLPs when implemented on FPGA-based systems, particularly in high-dimensional classification tasks. While KANs may offer comparable accuracy for simpler datasets, their resource consumption renders them inefficient for complex, large-scale applications. 

To address some of these challenges, Liu et al. \cite{liu2024kan20kolmogorovarnoldnetworks} proposed KAN 2.0, which introduced the MultKAN model with multiplication nodes to reduce computational complexity and improve interpretability. However, while these enhancements help mitigate some of the training time issues, KANs still exhibit longer training times and higher resource usage compared to other architectures due to their reliance on symbolic regression and specialized activation functions. In the context of dynamical systems, Koenig et al. \cite{koenig2024kan} demonstrated the application of KAN-ODEs (Kolmogorov-Arnold Network Ordinary Differential Equations), which outperform traditional Neural ODEs in terms of accuracy and interpretability. Yet, the study acknowledges that KANs require specialized optimization techniques to balance these benefits, which introduces additional computational overhead.

Furthermore, recent advancements in rKAN by Aghaei \cite{aghaei2024rkan}  show promise in improving computational efficiency through rational functions for smoother function approximation. However, these improvements come at the cost of increased complexity during the design and training phases, especially in high-dimensional spaces. Schmidt-Hieber \cite{schmidt2021kolmogorov} revisited the Kolmogorov-Arnold representation theorem, noting that while KANs theoretically excel in decomposing high-dimensional functions, they face significant challenges related to scalability. In particular, the increased number of parameters and complex activation functions required for high-dimensional tasks result in greater computational resource consumption compared to deep ReLU networks.

\subsection{Generalization}
KANs have emerged as a promising alternative to traditional neural network architectures like MLPs and CNNs, particularly in scientific tasks and structured data applications. Recent studies have demonstrated that KANs offer competitive performance in terms of generalization across different datasets and problem domains. Zhang and Zhou \cite{zhang2024generalization} provide a comprehensive analysis of KANs' generalization abilities, establishing theoretical generalization bounds for networks equipped with activation functions based on basis functions or low-rank Reproducing Kernel Hilbert Space (RKHS). These bounds scale with operator norms, ensuring adaptability to varying complexities while offering empirical support for their effectiveness across both simulated and real-world datasets. 

Recent empirical studies have further explored how KANs perform in various contexts. Alter et al. \cite{alter2024robustness} analyzed the robustness of KANs under adversarial attacks, showing that KANs' decomposition into univariate components provides advantages in resisting adversarial perturbations, although they may still face overfitting challenges on smaller models. Techniques such as adversarial training, regularization methods like dropout and weight decay, and defensive distillation have been shown to improve KANs' robustness against overfitting and underfitting, particularly when compared to MLPs and CNNs.

In dynamical systems, Koenig et al. \cite{koenig2024kan} demonstrated that KANs can generalize effectively across diverse datasets, including time-series and scientific applications, by leveraging sparse regularization and symbolic constraints. These techniques not only enhance robustness but also reduce parameter count, outperforming Neural ODEs and other architectures in tasks requiring interpretability and computational efficiency. Similarly, Carlo et al. \cite{de2024kolmogorov} extended the Kolmogorov–Arnold Network concept to GNNs, showing that GKANs excel in node classification and link prediction tasks while providing superior interpretability. By incorporating spline-based activation functions and applying sparsification and pruning, GKANs address overfitting more effectively than traditional GNN models, allowing them to generalize across graph-structured data with minimal risk of underfitting.

Moreover, Samadi et al. \cite{samadi2024smooth} focused on the smoothness and structural knowledge in KANs, revealing that smooth KANs embedded with domain-specific knowledge can reduce the amount of data needed for training, thereby minimizing the risk of both overfitting and underfitting. This structural embedding allows KANs to generalize well across sparse datasets and improves convergence compared to standard MLPs. 

To address issues of overfitting and underfitting, several techniques have been applied to enhance KANs' robustness. Sparse regularization, entropy regularization, and grid extension techniques are commonly employed to prevent overfitting while improving interpretability and performance on smaller datasets \cite{liu2024kan, liu2024kan20kolmogorovarnoldnetworks}. Liu et al. \cite{liu2024kan20kolmogorovarnoldnetworks} introduced MultKAN, an augmented version of KANs incorporating multiplicative layers, which not only enhances model capacity but also improves interpretability by uncovering modular structures in the data. Compared to MLPs and CNNs, KANs excel in small-scale tasks due to their spline-based architecture, allowing for precise function fitting without excessive model complexity. Empirical studies demonstrate that KANs often outperform traditional architectures, particularly in low-dimensional settings, graph learning, and scientific discovery, where their ability to model compositional and univariate structures proves advantageous \cite{de2024kolmogorov}.

\subsection{Lack of Interpretability}
KANs have recently emerged as a promising alternative to traditional neural networks, offering enhanced interpretability by employing spline-parameterized functions instead of fixed weights on network edges. This architecture, grounded in the Kolmogorov-Arnold representation theorem, enables KANs to approximate complex, multivariate functions while maintaining a degree of transparency often absent in standard models like MLPs \cite{xu2024kolmogorov, galitsky2024kolmogorov, sun2024evaluating}. However, despite their interpretability advantages, KANs still encounter substantial challenges when compared to inherently interpretable models, such as decision trees and linear regression, which provide straightforward, rule-based insights directly aligned with feature inputs.

The interpretability challenges for KANs primarily stem from the complexity of their function compositions, which can obscure the underlying relationships in high-dimensional or nonlinear data. This is particularly problematic in sensitive applications like healthcare or finance, where model transparency is crucial for decision-making. To mitigate these challenges, explainability techniques like SHapley Additive exPlanations (SHAP) and Local Interpretable Model-Agnostic Explanations (LIME) offer promising support. While KANs possess intrinsic interpretability features such as symbolic regression in time series forecasting and scientific discovery SHAP and LIME can augment these by pinpointing feature contributions at a more granular level, potentially bridging the gap between KANs and simpler, interpretable models \cite{de2024kolmogorov, bresson2024kagnns}.

Several recent studies propose specific advancements for improving KAN interpretability. Xu et al. \cite{xu2024kolmogorov} suggest that symbolic regression within KANs, applied in models like T-KAN and MT-KAN, can decode nonlinear relationships over time series, enhancing transparency yet still falling short of simpler models in user-friendliness. Similarly, Galitsky \cite{galitsky2024kolmogorov} applies KAN to word-level explanations in NLP, integrating inductive logic programming to make language representations more understandable. Despite these efforts, additional techniques such as SHAP could offer improved clarity, especially where feature importance needs direct, interpretable weighting. Other works, including Carlo et al. \cite{de2024kolmogorov}  and Sun \cite{sun2024evaluating}, examine KAN’s applications in GNNs and scientific discovery, respectively, where transparency remains an issue due to the complex, multi-layered compositions of KANs. Finally, studies on KAN variants like Wav-KAN by Bozorgasl and Chen \cite{bozorgasl2024wav} highlight the potential of wavelet-based KAN configurations for interpretability, although even these enhancements would benefit from complementary explainability techniques for application-specific insights.

The mathematical foundation of KANs adds interpretability challenges due to their complex transformations. Schmidt-Hieber \cite{schmidt2021kolmogorov} notes that while KANs are expressive, their non-intuitive functions can reduce interpretability compared to simpler models. Constraining these transformations for smoother outputs may help, but tools like SHAP offer immediate insight by decomposing complex outputs into understandable parts. In dynamical systems, Koenig et al. \cite{koenig2024kan} demonstrate that embedding KANs in neural ODEs enhances interpretability for modeling complex dynamics, though understanding outputs in non-linear systems remains difficult. Here, SHAP could clarify component contributions in high-stakes applications like physics-based models.

\section{Current Trends and Advancements}
\subsection{Recent Developments}
Recent advancements in KAN have significantly expanded their performance, scalability, and applicability across diverse domains, fueled by innovations in architecture, training, and hybrid methods. Originally designed to leverage the Kolmogorov-Arnold representation theorem, KANs offer a unique structure that places activation functions on edges rather than nodes, allowing more flexible, modular, and efficient function approximations. In graph learning, KAN variants such as the KAGCN and KAGIN have demonstrated that KANs can outperform MLPs in specific graph tasks, notably in graph regression, by providing improved node feature updates \cite{bresson2024kagnns}. Meanwhile, in transfer learning, Shen and Younes \cite{shen2024reimagining} replaced the traditional linear probing layer in ResNet-50 with a KAN layer, achieving higher adaptability to complex data patterns and significantly enhancing model generalization. Additionally, the Residual KAN (RKAN), which incorporates KAN modules within deep CNNs using Chebyshev polynomial-based convolutions, has demonstrated that this hybrid approach can enhance feature extraction in ResNet and DenseNet architectures while retaining computational efficiency \cite{yu2024residual}.

KAN autoencoders also show promise, as they achieve competitive reconstruction accuracy in comparison to CNN autoencoders on benchmark datasets like MNIST, thanks to the edge-based activation functions that allow KANs to capture nuanced data dependencies, promoting their use in data representation tasks \cite{moradi2024kolmogorov}. Despite these strengths, KANs are sensitive to noise, which can degrade performance. To address this, Shen et al. \cite{shen2024reduced} proposed kernel filtering and oversampling techniques, improving KAN robustness against noisy datasets, which is crucial for their applicability in real-world data environments. In time series forecasting, KANs have demonstrated notable efficiency and predictive accuracy, particularly with satellite traffic data, showing potential for broader applications in sequential data analysis by dynamically learning activation patterns \cite{vaca2024kolmogorov}.

\begin{figure}[htp]
    \centering
    \includegraphics[width=0.98\linewidth]{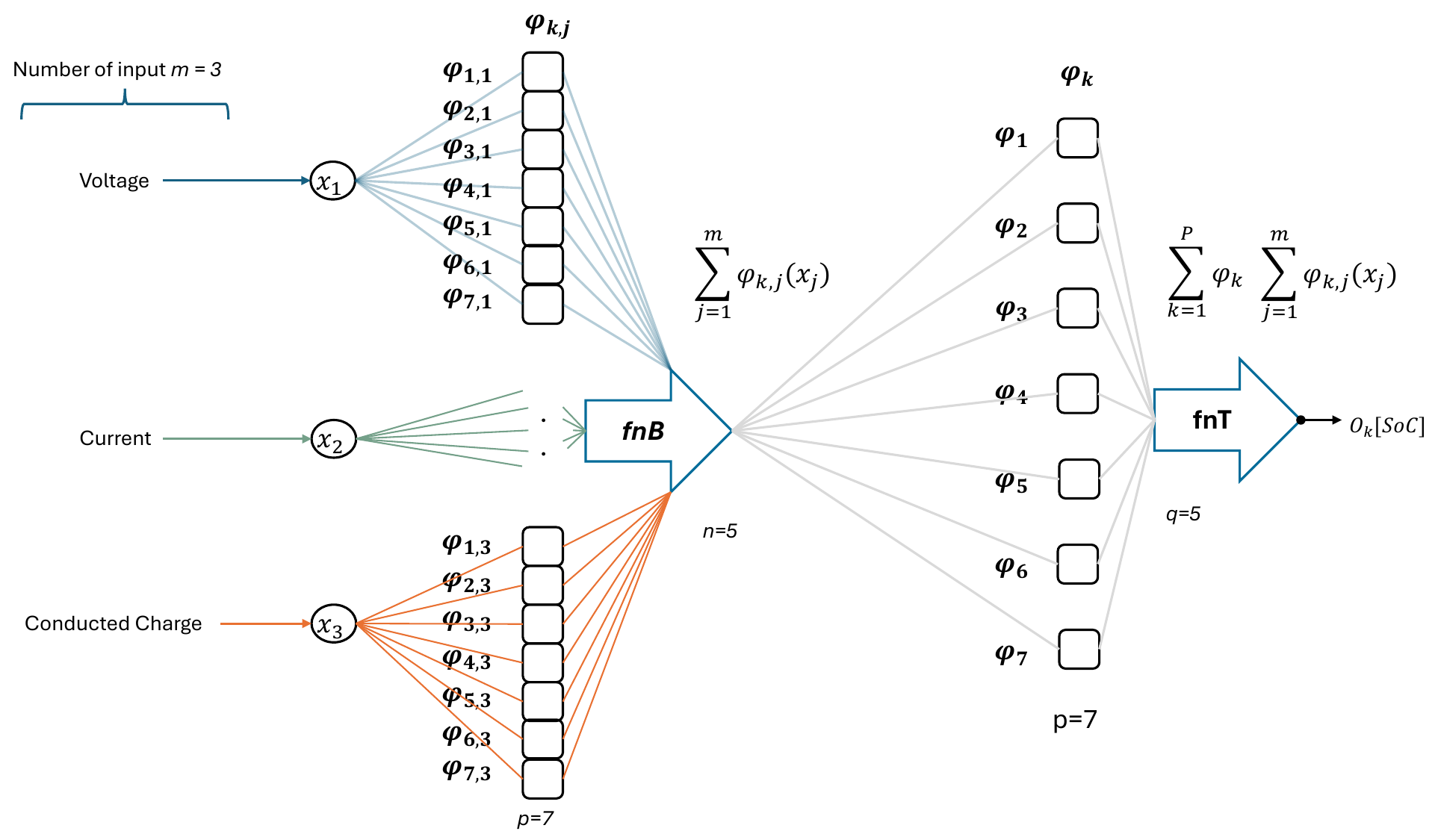}
    \caption{Kolmogorov-Arnold Network Architecture for Estimating Battery State of Charge (SoC) \cite{sulaiman2024battery}}
    \label{fig:KANSoC}
\end{figure}

Moreover, recent applications of KAN in electric vehicle battery charge estimation have highlighted their scalability and precision in handling high-dimensional data, surpassing ANN and a hybrid Barnacles Mating Optimizer-deep learning model through KAN’s high-dimensional adaptability \cite{sulaiman2024battery}. As illustrated in Figure \ref{fig:KANSoC}, the network architecture for SoC estimation leverages KAN’s edge-based activation functions to process complex inputs like voltage, current, and conducted charge, enhancing the model’s accuracy in nonlinear scenarios. Another notable application in chiller energy consumption prediction for commercial buildings showcases KAN’s ability to model complex, nonlinear dynamics effectively. Compared against both ANN and hybrid deep learning models, KAN demonstrates superior accuracy and computational efficiency, confirming its role as a viable option for optimizing energy management \cite{sulaiman2024utilizing}.

The smooth KAN (SKAN) architecture developed by Samadi et al. integrates structural knowledge to improve interpretability and reduce data requirements. This variant, focusing on computational biomedicine, addresses KAN's convergence limitations and provides a reliable approach for tasks that benefit from prior structural insights \cite{samadi2024smooth}. In hyperspectral image classification, a hybrid KAN architecture with 1D, 2D, and 3D modules is shown to outperform CNNs and vision transformer models by enhancing feature discrimination in low-dimensional data, an advancement that proves valuable in remote sensing and Earth observation tasks \cite{jamali2024learn}.

In adversarial robustness research, Alter et al. \cite{alter2024robustness} illustrate that KANs have lower Lipschitz constants compared to MLPs, providing greater resistance to perturbations in adversarial conditions. This robustness marks KAN as a promising model for security-sensitive applications. Finally, Poeta et al. \cite{poeta2024benchmarking} benchmarking study on real-world tabular datasets indicates that KAN excels in accuracy and interpretability but comes with a higher computational cost. This finding suggests KAN's suitability as an MLP alternative for complex, large-scale tabular data, further expanding its scope in practical machine learning applications.

\subsection{Integration with Other Models}
To explore the benefits of integrating KAN with models like GNNs, CNNs, Recurrent Neural Networks (RNNs), and Reinforcement Learning, recent studies have extended KAN's structure across these architectures. KANs, grounded in the Kolmogorov–Arnold representation theorem, compose models from simpler, interpretable functions rather than dense weight matrices, promoting transparency. This structure is especially valuable in fields requiring interpretability, including graph data analysis, molecular property prediction, and network classification \cite{de2024kolmogorov, li2024ka}.

Recent studies have explored the combination of KAN with convolutional layers, primarily for applications in image classification and time-series forecasting. For instance, Convolutional KANs (C-KAN) introduce convolutional layers that enhance KAN’s feature extraction capability for complex, high-dimensional data such as images and time series, showing improved prediction accuracy and resilience to non-stationarity in data but without extending to GNNs, RNNs, or Reinforcement Learning applications \cite{bodner2024convolutional, livieris2024c}. KAN with Interactive Convolutional Elements (KANICE), a KAN architecture incorporating Interactive Convolutional Blocks (ICBs), also aims to improve CNN adaptability by dynamically adjusting feature extraction across varying input distributions. While highly effective for image classification, this approach has not yet extended to domain-specific tasks like graph analysis or sequential processing \cite{ferdaus2024kanice}. Additionally, Activation Space Selectable KAN (S-KAN) introduces selectable activation modes to increase model adaptability across general data-fitting tasks and standard image classification datasets, showing promising results yet without addressing integrations with GNNs or sequential models such as RNNs \cite{yang2024activation}.  

In physics-informed contexts, KANs have been adapted to Physics-Informed Neural Networks (PINNs), creating Kolmogorov-Arnold-Informed Networks (KINNs) for solving PDEs with greater accuracy than traditional neural networks. This application leverages KAN’s spline-based architecture to improve convergence speed and parameter efficiency, beneficial in computational physics tasks, though it remains limited to PDE-focused applications and does not integrate GNN or sequential processing capabilities \cite{wang2024kolmogorov, yu2024sinc}. Further extending KAN's framework, Wang et al. \cite{wang2024kolmogorov} explore its integration within the energy form of PDEs, enabling KANs to enhance traditional PINN architectures without exploring graph data or real-time applications like reinforcement learning.

Studies, such as by Kilani \cite{kilani2024kolmogorov}, have underscored the versatility of KANs, highlighting successful applications in temporal data analysis and multi-step time series forecasting through hybrid KAN-RNN architectures, albeit on a foundational level. Integrations with advanced reinforcement learning or specialized GNN architectures have not been explored, suggesting the potential for KAN frameworks to enhance model interpretability and parameter efficiency across these domains in future work. In summary, current research indicates that the integration of KAN with other neural network architectures like CNNs has led to parameter-efficient, adaptable models suited to high-dimensional applications in image processing and time-series analysis. However, significant gaps remain in exploring KAN’s potential with GNNs, Reinforcement Learning, and RNNs for domain-specific tasks such as graph data analysis and sequential decision-making.

\section{Future Directions}
KANs present substantial opportunities and challenges across diverse fields, yet key limitations impact their scalability, benchmarking, and interdisciplinary applications. One significant limitation is scalability, particularly due to KAN's complex, spline-based architecture, which, while advantageous for interpretability, increases computational demands and training time in high-dimensional environments. Addressing scalability issues could enhance KAN’s application across fields such as environmental science, healthcare, and finance, where real-time and large-scale data processing is essential. Another critical area is benchmarking. KAN's unique structure lacks standardized benchmarking, limiting comparisons across fields like finance or healthcare where accuracy, speed, and interpretability are vital. Developing robust benchmarks for KANs in complex and noisy environments could facilitate interdisciplinary adoption, leading to breakthroughs in applications like ecological modeling, medical diagnostics, and risk management in finance. 

Enhancing KAN’s efficiency, adaptability, and generalization can be achieved through architectural innovations, optimization techniques, and training strategies. Architecturally, integrating KAN with models such as CNNs and RNNs, or exploring hybrid KAN variants for handling real-time data, could enable better feature extraction and adaptability in high-dimensional contexts. Innovations such as modular layers and residual connections could improve model efficiency, while advanced optimizations like adaptive gradient clipping and entropy-based regularization could mitigate overfitting, especially in noisy data environments. Further, training strategies incorporating multi-task learning and batch normalization may improve KAN's generalization in dynamic environments, which is crucial for real-time applications in fields like satellite monitoring and healthcare forecasting.

\section{Conclusion}
Kolmogorov-Arnold Networks (KANs) stand out as a promising and theoretically grounded alternative to conventional neural networks, leveraging the Kolmogorov-Arnold representation theorem to decompose high-dimensional, multivariate functions into simpler, univariate components. This review highlighted KANs’ unique strengths, including learnable spline-based activation functions, which allow precise function approximation with fewer parameters, enhancing interpretability and efficiency in applications such as time-series forecasting, graph learning, and physics-informed modeling. Recent advancements, including models like T-KAN and Wav-KAN, have showcased KAN’s adaptability to dynamic and spectral data, demonstrating competitive or superior performance in efficiency and predictive power over traditional models.

Despite these strengths, KANs face challenges in scalability, computational complexity, and noise sensitivity in high-dimensional environments. Integrating KAN with other architectures, such as CNNs, RNNs, and GNNs, is explored as a potential avenue to enhance flexibility and address scalability limitations. Such hybrid architectures could allow KANs to inherit the interpretability benefits of their design while leveraging the flexibility and efficiency of other models, making them viable for applications across domains with high data demands.

This systematic review underscores KAN’s growing relevance in scientific and applied research, proposing future directions that focus on optimizing scalability, improving noise robustness, and developing efficient training strategies. These advancements could strengthen KAN’s role in data-driven applications, contributing to the ongoing development of transparent, efficient neural networks well-suited for complex function approximation in various fields.

Future research could focus on optimizing KAN’s computational efficiency, exploring hybrid architectures with other models (e.g., CNNs, RNNs, GNNs), and developing robust benchmarks to facilitate interdisciplinary applications. Addressing these challenges could position KANs as a powerful tool for transparent and scalable neural networks in complex, data-driven domains.

\begin{acks}
We extend our sincere gratitude to Khaled Aly Abousabaa for his assistance in preparing the images for this paper. We also thank our colleagues at Texas State University for their valuable guidance throughout this work.
\end{acks}


\appendix

\end{document}